%% file: arxiv_main.tex
\documentclass[letterpaper]{article}

\usepackage{times}
\usepackage{helvet}
\usepackage{courier}
\usepackage{arxiv}
\usepackage{verbatim}

\usepackage[utf8]{inputenc} 
\usepackage[T1]{fontenc}    
\usepackage{hyperref}       
\usepackage{url}            
\usepackage{booktabs}       

\usepackage{subfig}
\usepackage{nicefrac}       
\usepackage{microtype}      
\usepackage{graphicx}
\usepackage{natbib}
\usepackage{doi}

\usepackage{multirow, rotating}

\usepackage{amsmath, amsthm, amssymb}
\usepackage{geometry}
\usepackage{booktabs}
\usepackage{array}
\usepackage{xcolor}
\usepackage{enumitem}
\usepackage{tikz}
\usepackage{algorithm}
\usepackage{algpseudocode}
\usepackage{caption}
\usetikzlibrary{positioning, arrows.meta, calc, shapes.geometric}

\hypersetup{
  colorlinks = true,
  linkcolor  = blue!60!black,
  citecolor  = blue!60!black,
  urlcolor   = blue!60!black
}

\theoremstyle{plain}

\newtheorem{theorem}{Theorem}[section]

\newtheorem{conjecture}[theorem]{Conjecture}

\theoremstyle{definition}
\newtheorem{definition}[theorem]{Definition}

\theoremstyle{remark}
\newtheorem{remark}[theorem]{Remark}

\title{Instruction Set and Language for Symbolic Regression}

\author{ \href{https://orcid.org/0000-0001-8231-5687}{\includegraphics[scale=0.06]{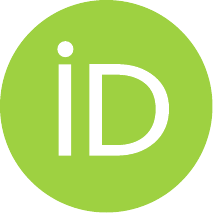}\hspace{1mm}Ezequiel L\'opez-Rubio}\thanks{Corresponding author. ITIS Software. Universidad de M\'alaga. C/ Arquitecto Francisco Peñalosa 18, 29010, Málaga, Spain} \\
	Department of Computer Languages and Computer Science\\
    University of M\'alaga\\
    Bulevar Louis Pasteur, 35\\
    29071 M\'alaga, Spain \\
	\texttt{ezeqlr@lcc.uma.es} \\
	\And
	\href{https://orcid.org/0009-0001-2178-4647}{\includegraphics[scale=0.06]{orcid.pdf}\hspace{1mm}Mario Pascual-Gonz\'alez} \\
    Department of Computer Languages and Computer Science\\
    University of M\'alaga\\
    Bulevar Louis Pasteur, 35\\
    29071 M\'alaga, Spain \\
	\texttt{mpascual@uma.es} \\
}

\renewcommand{\shorttitle}{Instruction Set and Language for Symbolic Regression}

\hypersetup{
pdftitle={Instruction Set and Language for Symbolic Regression},
pdfsubject={cs.AI, cs.CL},
pdfauthor={Ezequiel L\'opez-Rubio, Mario Pascual-Gonz\'alez},
pdfkeywords={}
}

\begin{document}
\maketitle

\begin{abstract}
A fundamental but largely unaddressed obstacle in Symbolic regression (SR) is structural redundancy:
every expression DAG with $k$ internal nodes admits $\Theta(k!)$ distinct
node-numbering schemes that all encode the \emph{same} expression, each
occupying a separate point in the search space and consuming fitness
evaluations without adding diversity.

We present \textsc{IsalSR} (Instruction Set and Language for Symbolic
Regression), a representation framework that encodes expression DAGs as
strings over a compact two-tier alphabet and computes a \emph{pruned
canonical string}---a complete labeled-DAG isomorphism invariant---that
collapses all $\Theta(k!)$ equivalent representations into a single canonical
form.

We validate five fundamental properties of the representation empirically on
the Nguyen and AI~Feynman benchmark suites: round-trip fidelity, DAG
acyclicity, canonical invariance and idempotence, evaluation preservation, and
search-space reduction.

\end{abstract}

\keywords{symbolic regression, canonical string representation, labeled directed acyclic graph, isomorphism invariant, search space reduction}

\section{Introduction\label{sec:introduction}}

\input{introduction}

\section{Methodology\label{sec:methodology}}

\input{methodology}

\section{Computational experiments\label{sec:Computational-experiments}}

\input{mario/computational_experiments.tex}
\section{Results\label{sec:results}}

\input{mario/results.tex}
\section{Conclusion\label{sec:conclusion}}

We have presented \textsc{IsalSR}, a representation framework that encodes
symbolic-regression expressions as instruction strings over a two-tier
alphabet and computes a pruned canonical string that serves as a complete
isomorphism invariant for labeled expression DAGs.
The framework addresses a structural redundancy inherent to all existing SR
methods: for a DAG with $k$ internal nodes, $\Theta(k!)$ distinct
node-numbering schemes encode the same expression and consume search budget
without contributing diversity.
\textsc{IsalSR} collapses this redundancy into a single canonical form per
isomorphism class and can be integrated into any existing SR algorithm as a
preprocessing step, requiring only a canonicalization call before fitness
evaluation.

The key technical contributions are threefold.
First, the two-tier alphabet guarantees that every string over the alphabet
decodes to a valid DAG, eliminating the need for grammar-based or
type-system-based validity filters that prune the search space through
rejection.
Second, the commutative encoding---replacing subtraction and division with
unary \textsc{Neg} and \textsc{Inv} operators---reduces the set of
non-commutative binary operations to \textsc{Pow} alone, simplifying the
isomorphism definition while preserving the full expressive power of the
standard SR function set.
Third, the label-aware pruned backtracking canonicalization algorithm
computes the canonical string efficiently, with scalability results to be
reported in a subsequent revision.

Empirical validation on the Nguyen and AI~Feynman benchmark suites
confirms the five fundamental properties of the representation.
Round-trip fidelity holds at 100\% across all tested expressions, and the
pruned algorithm agrees with the full canonical algorithm on more than 99.97\%
of test cases while producing strings up to 21\% shorter than the greedy
baseline.
Canonical invariance and idempotence are verified at 100\% across 961 DAGs,
supporting Conjecture~\ref{thm:invariant} and providing strong empirical
evidence that the pruned canonical string is a complete labeled-DAG invariant.
Evaluation preservation holds within a tolerance of $10^{-8}$ at every test
point.
Search-space analysis confirms that canonicalization collapses all
$\Theta(k!)$ equivalent node-numbering representations into a single
canonical form.
The Levenshtein distance on canonical strings induces a meaningful metric on
the space of expression isomorphism classes: single-character edits correspond
to semantically local changes such as operator substitution, and distance
values align with the structural dissimilarity between expression pairs.
Neighbourhood analysis around a representative expression reveals an overall
redundancy rate of 71.2\%, quantifying the search-space compression that
canonicalization provides.

\paragraph{Limitations.}
The formal proofs of the two central conjectures---round-trip fidelity
(Conjecture~\ref{thm:roundtrip}) and canonical invariance
(Conjecture~\ref{thm:invariant})---remain open and are left as future work.
The scalability study measuring CPU time of the pruned canonicalization
algorithm as a function of DAG size is currently under analysis and will be
reported in a subsequent revision.
Integration experiments with specific SR algorithms, such as GraphDSR
or BINGO, demonstrating end-to-end search-space reduction in terms of
fitness evaluations and solution quality, are likewise deferred to
future work.
Finally, the current alphabet is fixed to the function set of
Table~\ref{tab:operations}; extending \textsc{IsalSR} to user-defined
operator sets, conditional expressions, or multi-output DAGs requires
further investigation.

\paragraph{Future directions.}
Immediate priorities include formal proofs of the two conjectures, a
complete scalability study, and integration experiments with state-of-the-art
SR solvers.
Longer-term directions include exploiting the canonical string as a
deduplication key in population-based methods---replacing fitness re-evaluation
of isomorphic individuals---and as a structured representation for
language-model-based SR approaches, where the alphabet provides a natural
token vocabulary with a built-in validity guarantee.
The metric structure induced by the Levenshtein distance may also enable
diversity-preserving operators in genetic programming that explicitly
navigate the canonical string space.

\section*{Acknowledgment}
The authors thankfully acknowledge the computer resources (Picasso
Supercomputer), technical expertise, and assistance provided by the
SCBI (Supercomputing and Bioinformatics) center of the University
of M\'alaga.

\bibliographystyle{unsrtnat}
\bibliography{references}

\end{document}

%% file: introduction.tex
Symbolic regression (SR) is the task of discovering a closed-form
mathematical expression $f(x_1,\ldots,x_m)$ that fits an observed dataset
$(X,y)$, without restricting the functional form \emph{a priori}.
Unlike parametric regression, SR simultaneously searches over the space of
mathematical structures and their numerical parameters, making it
combinatorially challenging but uniquely interpretable: a successful model
is not a black box but a human-readable formula that can be examined,
simplified, and extrapolated beyond the training range~\citep{koza1992}.
This interpretability has driven applications across scientific
discovery~\citep{udrescu2020}, engineering~\citep{randall2022bingo},
and machine learning~\citep{lacava2021}.

Despite decades of progress, SR search spaces suffer from a structural
redundancy that no existing method addresses directly.
A mathematical expression can be represented as a labeled Directed Acyclic
Graph (DAG) in which leaf nodes are input variables or learnable constants,
and internal nodes are arithmetic or transcendental operations.
For a DAG with $k$ internal nodes, there exist $\Theta(k!)$ distinct
node-numbering schemes that all encode the \emph{same} expression.
Every one of these equivalent representations occupies a separate point in
the search space, consuming fitness evaluations and population slots without
adding diversity.
For $k=8$ internal nodes---a modest expression size---this amounts to over
$40{,}000$ redundant copies of every structurally unique candidate.
No current SR method deduplicates at this granularity.

\medskip
\noindent\textbf{Contribution.}
We present \textsc{IsalSR} (Instruction Set and Language for Symbolic
Regression), a representation framework that encodes expression DAGs as
instruction strings and computes a \emph{pruned canonical string}---a
complete labeled-DAG invariant---that collapses all $\Theta(k!)$ equivalent
representations into a single canonical form.
Any existing SR algorithm can be augmented with \textsc{IsalSR} by
inserting a single canonicalization step before fitness evaluation,
reducing the effective search space by $O(k!)$ without altering the
algorithm's core logic.

\textsc{IsalSR} shares some similarities with IsalGraph~\citep{lopezrubio2025}, a prior framework
for encoding unlabeled undirected graphs as instruction strings. However, \textsc{IsalSR} is fundamentally different because it is devoted to the labeled, directed, acyclic setting required for symbolic
regression.
The key contributions of \textsc{IsalSR} are: (i) a two-tier instruction alphabet that
encodes both graph topology and operation labels so that all strings over the alphabet represent a valid DAG; (ii) a commutative
encoding that replaces binary non-commutative operators (subtraction,
division) with unary decomposition operators, simplifying the isomorphism
definition; (iii) a label-aware pruned backtracking algorithm for computing
the canonical string efficiently; and (iv) the formal property that the canonical
string is a complete isomorphism invariant for labeled DAGs under the
adopted semantics, including operand-order preservation for the sole
remaining non-commutative operator (\textsc{Pow}).

\subsection{Related Work}
\label{sec:related}

\paragraph{Genetic programming.}
The dominant paradigm for SR since Koza~\citep{koza1992} is genetic
programming (GP), which evolves populations of expression trees under
crossover and mutation.
GP naturally produces redundant populations: two individuals with the same
expression but different tree shapes are treated as distinct.
Semantic crossover operators~\citep{uy2011} partially address this by
considering behavioral equivalence, but do not handle structural
isomorphism.
Canonicalization is orthogonal to and composable with all GP variants.

\paragraph{Deep learning and reinforcement learning approaches.}
Petersen et~al.~\citep{petersen2021} propose Deep Symbolic Regression (DSR),
which trains an RNN with a risk-seeking policy gradient to generate
expression trees sequentially.
The recurrent structure again lacks any mechanism for detecting that two
generated sequences correspond to the same expression.
Liu et~al.~\citep{liu2025} introduce GraphDSR, which represents expressions
as DAGs rather than trees, enabling shared subexpression reuse and achieving
state-of-the-art performance on the Nguyen and AI~Feynman benchmarks.
GraphDSR's DAG representation makes the redundancy problem more acute:
variable-arity addition and multiplication nodes admit many equivalent
orderings of their inputs, all of which a DAG-based method may explore
without deduplication.
\textsc{IsalSR}'s canonical string directly resolves this ambiguity for
GraphDSR-style expression DAGs.

\paragraph{Graph-based symbolic regression.}
Xiang et~al.~\citep{xiang2025graphsr} observe, in the context of GraphSR,
that non-commutative binary operations (subtraction, division) introduce
operand-order ambiguity that complicates the isomorphism definition.
They propose a unary decomposition: $x-y = \mathrm{Add}(x, \mathrm{Neg}(y))$
and $x/y = \mathrm{Mul}(x, \mathrm{Inv}(y))$, replacing binary
non-commutative nodes with commutative binary nodes and new unary operators.
\textsc{IsalSR} adopts this decomposition, reducing the set of
non-commutative binary operations to \textsc{Pow} alone.

\paragraph{Uniform depth-first search (UDFS).}
Kahlmeyer et~al.~\citep{kahlmeyer2024udfs} propose a structured enumeration
strategy for SR based on uniform depth-first search over expression DAGs,
with topological numbering to generate candidates systematically.
UDFS demonstrates that structured traversal of the expression space
improves coverage over random search, but it does not identify or collapse
structurally isomorphic DAGs.
The canonical string of \textsc{IsalSR} provides a natural deduplication
key that can be integrated into UDFS-style enumeration: candidates with
the same canonical string are isomorphic and can be skipped.

\paragraph{Physics-informed and grammar-based SR.}
Udrescu and Tegmark~\citep{udrescu2020} leverage physical symmetry groups
and dimensional analysis to constrain the search space of the AI~Feynman
benchmark.
Grammar-guided GP methods impose syntactic constraints via context-free
grammars to prune invalid expressions.
These approaches reduce the search space by restricting which expressions
are generated, whereas \textsc{IsalSR} has no invalid expressions without any need of pruning because each string over the defined alphabet of instructions is associated with a valid DAG. Moreover, \textsc{IsalSR} reduces the search space even further by identifying which
generated strings are equivalent.

\paragraph{BINGO.}
Randall et~al.~\citep{randall2022bingo} introduce BINGO (Boosting of
Implicit Non-linear Gene expression programming with Optimization), an
open-source SR tool developed at NASA for engineering and scientific
applications.
BINGO combines age-fitness Pareto optimization with automatic
differentiation and constant optimization via BFGS, achieving competitive
performance on aerospace applications.
Like all GP-based methods, BINGO's population-based search is susceptible
to the $O(k!)$ structural redundancy that \textsc{IsalSR} eliminates;
its efficient constant optimizer and modular architecture make it a natural
host for \textsc{IsalSR} canonicalization.

\paragraph{SRBench.}
La~Cava et~al.~\citep{lacava2021} provide a large-scale benchmark comparing
contemporary SR methods on 252~real-world regression datasets drawn from
PMLB, covering methods including GP, DSR, and linear baselines.
Their comprehensive evaluation reveals that no single method dominates
across all datasets, motivating orthogonal improvements such as the
search-space reduction provided by \textsc{IsalSR}.

\subsection{Paper Organization}


The remainder of the paper is structured as follows.
Section~\ref{sec:methodology} defines the \textsc{IsalSR} instruction set,
the String-to-DAG and DAG-to-String algorithms, the pruned canonical string,
and the isomorphism invariance property.
Section~\ref{sec:benchmarks} describes the experimental design, including
the Nguyen, AI~Feynman, and SRBench benchmark suites.
Section~\ref{sec:results} presents empirical validation of the five
fundamental properties of the canonical representation, as well as empirical proof regarding the representation collapse of isomorphic DAGs to a single instruction string (Section~\ref{sec:results_search_space}).
Section~\ref{sec:conclusion} summarizes contributions and outlines future
directions.

%% file: methodology.tex
This section presents the formal foundations of \textsc{IsalSR}.
We define the instruction set~(\S\ref{sec:iset}), the String-to-DAG
decoder~(\S\ref{sec:s2d}), the DAG-to-String encoder~(\S\ref{sec:d2s}),
the pruned canonical string~(\S\ref{sec:canonical}), and conjectures that it is
a complete graph invariant under labeled-DAG isomorphism~(\S\ref{sec:invariant}).

\subsection{The IsalSR Instruction Set}
\label{sec:iset}

\textsc{IsalSR} represents symbolic expressions as \emph{labeled directed
acyclic graphs} (DAGs), encoded as strings over a compact two-tier alphabet.
Before defining the alphabet we fix the underlying data structure.

\begin{definition}[Labeled DAG]
\label{def:ldag}
A \emph{labeled DAG} is a tuple $D=(V,E,\ell,\delta)$ where:
\begin{enumerate}[nosep, label=(\roman*)]
  \item $V=\{0,1,\ldots,n{-}1\}$ is a finite node set with a fixed total
    order given by the integer indices;
  \item $E\subseteq V\times V$ is a set of directed edges such that $(V,E)$
    is acyclic;
  \item $\ell: V\to\mathcal{T}$ assigns each node an \emph{operation type}
    from the type set $\mathcal{T}$ (Table~\ref{tab:operations});
  \item $\delta: V\to\mathbb{Z}\cup\mathbb{R}$ stores optional per-node
    metadata: the variable index~$i$ for \textsc{Var} nodes ($\delta(v)=i$),
    the constant value for \textsc{Const} nodes, and zero for all other nodes.
\end{enumerate}
An edge $(u,v)\in E$ means ``$u$ provides input to $v$'' (data-flow direction).
The \emph{in-neighbors} of $v$ are $N^{-}(v)=\{u : (u,v)\in E\}$; the
\emph{out-neighbors} are $N^{+}(v)=\{u : (v,u)\in E\}$.
\end{definition}

\noindent\textbf{Operand order for \textsc{Pow}.}
Each node $v$ maintains an \emph{ordered input list}
$\sigma(v) = (s_1, s_2, \ldots)$, recording the in-neighbors of $v$ in the
order their connecting edges were added to the DAG.
\textsc{Pow} is the sole binary non-commutative operation: if
$\sigma(v)=(u_1, u_2)$, the evaluation convention is
\begin{equation}
  \label{eq:operand_order}
  \mathrm{eval}(v) = \mathrm{eval}(u_1)^{\,\mathrm{eval}(u_2)},
\end{equation}
where $u_1$ is the \emph{base} and $u_2$ is the \emph{exponent}.
The first edge added to $v$ determines $u_1$: the $\texttt{V}\ell$/$\texttt{v}\ell$ creation
token adds the creation edge first, so the creation-edge source is always
$\sigma(v)[0]$, the base.
A subsequent $C$/$c$ instruction appends the exponent $\sigma(v)[1]$.
All other operation nodes (\textsc{Add}, \textsc{Mul}, and all unary ops)
are commutative, so the order of their in-neighbors does not affect evaluation.

\noindent\textbf{Machine state.}
The \textsc{IsalSR} machine state is the quadruple $\mathcal{S}=(D,L,p,q)$,
where $D$ is the partially-built labeled DAG, $L$ is the Circular
Doubly-Linked List (CDLL) whose elements are node identifiers from $V$,
$p$ is the CDLL index of the primary pointer, and $q$ the secondary pointer.
The graph node at CDLL position $r$ is retrieved as $\mathtt{val}(r)\in V$. Conversely, the position within the CDLL of graph node $v$ is retrieved as $\mathtt{ptr}(v)\in V$.

\begin{table}[ht]
\centering
\caption{IsalSR operation types, label characters, arities, and evaluation
  semantics.
  Operations marked $\dagger$ use a numerically protected implementation.
  \textsc{Pow} is the only binary non-commutative operation: its operand
  order is given by $\sigma(v)$ (edge-insertion order), with $\sigma(v)[0]$
  as the base and $\sigma(v)[1]$ as the exponent.
  \textsc{Neg} and \textsc{Inv} are the unary decomposition operators that
  encode subtraction and division commutatively
  (see \S\ref{sec:commutative}).}
\label{tab:operations}
\begin{tabular}{lllp{5.2cm}}
\toprule
Label $\ell$ & Type & Arity & Evaluation \\
\midrule
$\texttt{+}$ & \textsc{Add}  & variadic (${\geq}2$) & $\sum_{u\in N^-(v)}\mathrm{eval}(u)$ \quad (commutative) \\
$\texttt{*}$ & \textsc{Mul}  & variadic (${\geq}2$) & $\prod_{u\in N^-(v)}\mathrm{eval}(u)$ \quad (commutative) \\
$\texttt{g}$ & \textsc{Neg}  & unary (1) & $-\mathrm{eval}(u)$ \\
$\texttt{i}$ & \textsc{Inv}  & unary (1) & $1\;/\;\mathrm{eval}(u)$ $\dagger$ \\
$\texttt{s}$ & \textsc{Sin}  & unary (1)  & $\sin(\mathrm{eval}(u))$ \\
$\texttt{c}$ & \textsc{Cos}  & unary (1)  & $\cos(\mathrm{eval}(u))$ \\
$\texttt{e}$ & \textsc{Exp}  & unary (1)  & $\exp(\mathrm{eval}(u))$ $\dagger$ \\
$\texttt{l}$ & \textsc{Log}  & unary (1)  & $\ln(\mathrm{eval}(u))$ $\dagger$ \\
$\texttt{r}$ & \textsc{Sqrt} & unary (1)  & $\sqrt{\mathrm{eval}(u)}$ $\dagger$ \\
$\texttt{a}$ & \textsc{Abs}  & unary (1)  & $|\mathrm{eval}(u)|$ \\
$\hat{\phantom{x}}$ & \textsc{Pow} & binary (2) & $\mathrm{eval}(\sigma(v)[0])^{\,\mathrm{eval}(\sigma(v)[1])}$ $\dagger$ \\
$\texttt{k}$ & \textsc{Const}& leaf (0)   & Learnable real constant \\
\midrule
(fixed) & \textsc{Var} & leaf (0) & Input variable $x_i$ (pre-inserted) \\
\bottomrule
\end{tabular}
\end{table}

\begin{definition}[IsalSR Instruction Set $\Sigma_{\mathrm{SR}}$]
\label{def:alphabet}
Let $\mathcal{L}=\{\texttt{+},\texttt{*},\texttt{-},\texttt{/},\texttt{s},\texttt{c},\texttt{e},\texttt{l},\texttt{r},\hat{\phantom{\texttt{x}}},\texttt{a},\texttt{k}\}$ be the set
of label characters.
The alphabet $\Sigma_{\mathrm{SR}}$ consists of four categories of tokens:
\begin{enumerate}[nosep]
  \item \textbf{Primary pointer movement} (single character):
    $\texttt{N}$ (advance to next CDLL element),
    $\texttt{P}$ (retreat to previous CDLL element).
  \item \textbf{Secondary pointer movement} (single character):
    $\texttt{n}$ (advance), $\texttt{p}$ (retreat).
  \item \textbf{Edge creation} (single character):
    $\texttt{C}$ creates the directed edge $\mathtt{val}(p)\to\mathtt{val}(q)$;
    $\texttt{c}$ creates $\mathtt{val}(q)\to\mathtt{val}(p)$.
    If the edge would create a cycle, it is silently skipped.
  \item \textbf{No-operation} (single character): $\texttt{W}$.
  \item \textbf{Labeled node insertion} (two-character token):
    $\texttt{V}\ell$ creates a new node $v_{\mathrm{new}}$ of type $\ell$,
    adds the directed edge $\mathtt{val}(p)\to v_{\mathrm{new}}$,
    and inserts $v_{\mathrm{new}}$ into the CDLL immediately after the
    primary-pointer position (the pointer does not move).
    $\texttt{v}\ell$ does the same using the secondary pointer $q$.
\end{enumerate}
The vocabulary contains $7$ single-character tokens and $24$ compound tokens
($2\times|\mathcal{L}|$), totaling $\mathbf{31}$ tokens.
\end{definition}

\noindent\textbf{Commutative encoding of subtraction and division.}
\label{sec:commutative}
The two unary operators \textsc{Neg} (label \texttt{g}, $-x$) and
\textsc{Inv} (label \texttt{i}, $1/x$) implement subtraction and division
through a commutative decomposition inspired by GraphSR~\citep{xiang2025graphsr}:
\[
  x - y \;=\; \textsc{Add}\!\bigl(x,\;\textsc{Neg}(y)\bigr),
  \qquad
  x / y \;=\; \textsc{Mul}\!\bigl(x,\;\textsc{Inv}(y)\bigr).
\]
Both \textsc{Add} and \textsc{Mul} are commutative, so the inputs to
these nodes are interchangeable and no operand-order tracking is required.
The protected inverse is $\textsc{Inv}(x) = 1/x$ when $|x| > \varepsilon$,
and $1$ otherwise.

\medskip
\noindent\textbf{Initial state.}
Given $m\geq 1$ input variables, the machine is initialized as:
\begin{itemize}[nosep]
  \item $V_0=\{x_1,\ldots,x_m\}$ with node IDs $0,\ldots,m{-}1$ assigned
    in variable-index order; $\ell(x_i)=\textsc{Var}$, $\delta(x_i)=i$.
  \item $E_0=\emptyset$.
  \item $L_0=[x_1,\ldots,x_m]$ (circular, in variable-index order).
  \item $p_0=q_0=\mathtt{ptr}(x_1)$.
\end{itemize}
Because node IDs are assigned in insertion order---starting from $0$ for
$x_1$---any node inserted later receives a strictly higher ID.
This convention makes the operand-order semantics of
Equation~\eqref{eq:operand_order} well-defined and deterministic for every
DAG produced by S2D.

\medskip
\noindent\textbf{Acyclicity invariant.}
After every token, $(V,E)$ remains acyclic.
For $\texttt{V}\ell$/$\texttt{v}\ell$, new nodes have no outgoing edges at creation, so
adding an edge \emph{to} them cannot create a cycle.
For $C$/$c$, a BFS reachability test is performed before insertion; the
instruction is silently skipped if a cycle would result.

\subsection{String-to-DAG Algorithm (S2D)}
\label{sec:s2d}

The \emph{String-to-DAG} (S2D) algorithm is a deterministic state machine
that decodes an IsalSR string $w\in\Sigma_{\mathrm{SR}}^*$ into a labeled DAG.
The LabeledDAG data structure tracks, for each node $v$, an ordered input
list $\sigma(v)$ that records in-neighbors in edge-insertion order.
For the sole binary non-commutative operation \textsc{Pow}, this list
encodes evaluation semantics via Equation~\eqref{eq:operand_order}: the
$\texttt{V}\ell$/$\texttt{v}\ell$ creation token establishes $\sigma(v)[0]$ (the base) by
adding the creation edge first; the subsequent $C$/$c$ token sets
$\sigma(v)[1]$ (the exponent).

\begin{definition}[S2D Execution]
\label{def:s2d}
Given $w\in\Sigma_{\mathrm{SR}}^*$ and $m\geq 1$, let
$(t_1,\ldots,t_k)=\mathtt{Tokenize}(w)$ be the token sequence produced by
a two-tier scanner (Definition~\ref{def:alphabet}).
Starting from $\mathcal{S}_0$, token $t_j$ induces the transition
$\mathcal{S}_{j-1}\to\mathcal{S}_j$ according to Table~\ref{tab:s2d_pseudo}.
The output is $\mathrm{S2D}(w,m) = D_k$.
\end{definition}

\begin{table}[ht]
\centering
\caption{Pseudocode for the String-to-DAG (S2D) algorithm.
  Each node $v$ maintains an ordered input list $\sigma(v)$; edges are
  appended to it in insertion order, establishing operand semantics for
  non-commutative binary nodes.}
\label{tab:s2d_pseudo}
\begin{tabular}{l}
\toprule
\textbf{Algorithm S2D}$(w,\,m)$ \\
\midrule
\textbf{Input:} instruction string $w$; number of variables $m \geq 1$ \\
\textbf{Output:} labeled DAG $D = (V, E, \ell, \delta, \sigma)$ \\[3pt]
\textit{// Initialization} \\
Create $m$ \textsc{Var} nodes $x_1,\ldots,x_m$ with IDs $0,\ldots,m{-}1$;
  \; set $E \leftarrow \emptyset$;\; $\sigma(v)\leftarrow[]$ for all $v$ \\
Insert $x_1,\ldots,x_m$ into CDLL $L$ in variable-index order \\
$p \leftarrow \mathtt{ptr}(x_1)$;\quad $q \leftarrow \mathtt{ptr}(x_1)$ \\[4pt]
\textbf{foreach} token $t \in \mathtt{Tokenize}(w)$ \textbf{do} \\[2pt]
\quad \textbf{if} $t = \texttt{N}$ \textbf{then} $p \leftarrow \mathtt{next}(L,p)$ \\
\quad \textbf{elif} $t = \texttt{P}$ \textbf{then} $p \leftarrow \mathtt{prev}(L,p)$ \\
\quad \textbf{elif} $t = \texttt{n}$ \textbf{then} $q \leftarrow \mathtt{next}(L,q)$ \\
\quad \textbf{elif} $t = \texttt{p}$ \textbf{then} $q \leftarrow \mathtt{prev}(L,q)$ \\
\quad \textbf{elif} $t = \texttt{C}$ \textbf{then} \\
\qquad $u \leftarrow \mathtt{val}(L,p)$;\; $v \leftarrow \mathtt{val}(L,q)$ \\
\qquad \textbf{if} $v$ not reachable from $u$ in $D$ \textbf{then} \\
\qquad\quad $E \leftarrow E \cup \{(u,v)\}$;\;
           $\sigma(v)\mathrel{+}=[u]$
  \hfill\textit{// append $u$ to $v$'s ordered input list} \\
\quad \textbf{elif} $t = \texttt{c}$ \textbf{then} \\
\qquad $u \leftarrow \mathtt{val}(L,q)$;\; $v \leftarrow \mathtt{val}(L,p)$ \\
\qquad \textbf{if} $v$ not reachable from $u$ in $D$ \textbf{then} \\
\qquad\quad $E \leftarrow E \cup \{(u,v)\}$;\;
           $\sigma(v)\mathrel{+}=[u]$ \\
\quad \textbf{elif} $t = \texttt{W}$ \textbf{then} \textit{(no-op)} \\
\quad \textbf{elif} $t = \texttt{V}\ell$ \textbf{then} \\
\qquad $u \leftarrow \mathtt{val}(L,p)$ \\
\qquad Create new node $v$ with type $\ell$ and next available ID; \;
       $V \leftarrow V \cup \{v\}$;\; $\sigma(v)\leftarrow[]$ \\
\qquad $E \leftarrow E \cup \{(u,v)\}$;\;
       $\sigma(v)\mathrel{+}=[u]$
  \hfill\textit{// $u$ becomes $\sigma(v)[0]$, the first operand} \\
\qquad $\mathtt{insert\_after}(L,\,p,\,v)$;\; $p$ unchanged \\
\quad \textbf{elif} $t = \texttt{v}\ell$ \textbf{then} \\
\qquad $u \leftarrow \mathtt{val}(L,q)$ \\
\qquad Create new node $v$ with type $\ell$ and next available ID; \;
       $V \leftarrow V \cup \{v\}$;\; $\sigma(v)\leftarrow[]$ \\
\qquad $E \leftarrow E \cup \{(u,v)\}$;\;
       $\sigma(v)\mathrel{+}=[u]$ \\
\qquad $\mathtt{insert\_after}(L,\,q,\,v)$;\; $q$ unchanged \\[4pt]
\textbf{return} $D = (V, E, \ell, \delta, \sigma)$ \\
\bottomrule
\end{tabular}
\end{table}

\begin{figure}[hbtp]
    \centering
    \includegraphics[width=\linewidth]{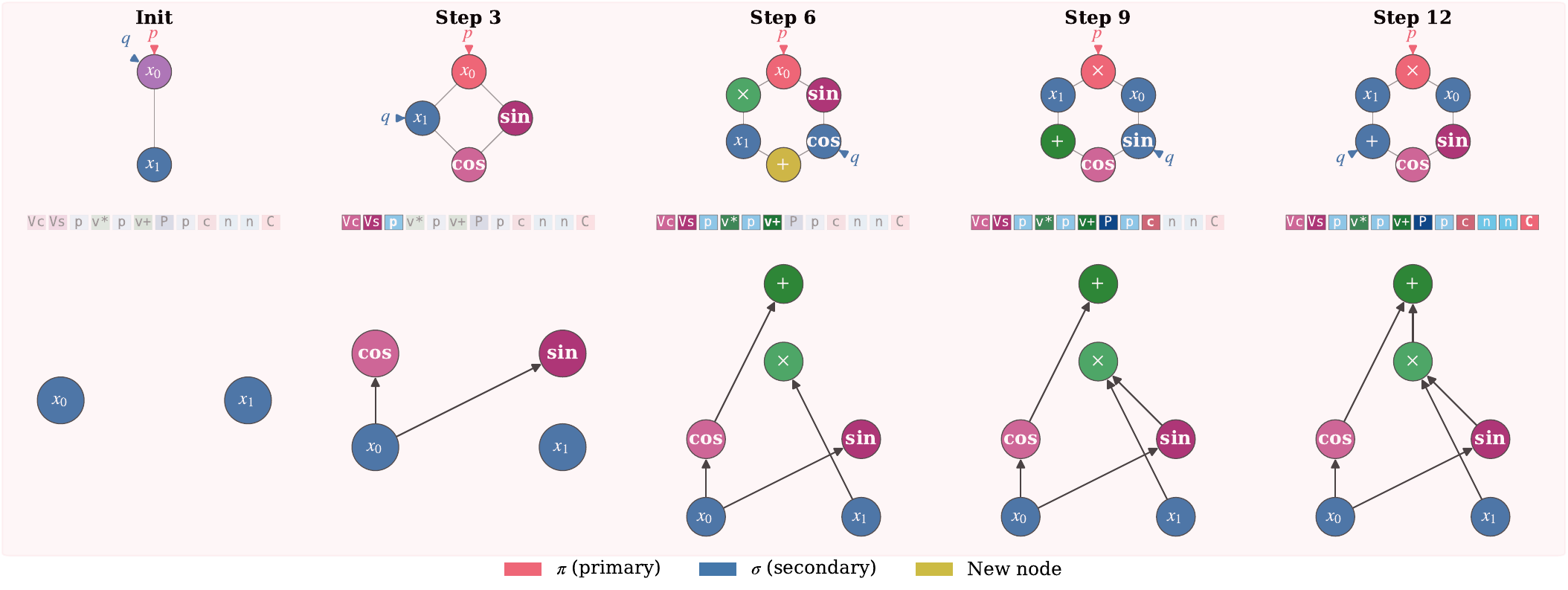}
    \caption{String-to-DAG execution for the canonical string \texttt{VcVspv*pv+PpcnnC}. The DAG grows
  incrementally as each instruction inserts labeled nodes (\texttt{V}/\texttt{v}) or directed edges (\texttt{C}/\texttt{c}), with
  primary ($\pi$) and secondary ($\sigma$) pointers navigating the circular doubly linked
  list.}
    \label{fig:s2d}
\end{figure}

\subsection{DAG-to-String Algorithm (D2S)}
\label{sec:d2s}

The \emph{DAG-to-String} (D2S) algorithm is a greedy encoder.
Given a labeled DAG $D$, it produces a string $w$ such that
$\mathrm{S2D}(w,m)\cong D$, where $\cong$ stands for Labeled-DAG isomorphism (see Definition \ref{def:isomorphism}).
D2S maintains a parallel \emph{output DAG} $D'$ alongside bidirectional
index maps $\mathtt{i2o}:V_D\!\to V_{D'}$ and $\mathtt{o2i}:V_{D'}\!\to V_D$,
initialized by mapping the $m$ \textsc{Var} nodes in variable-index order.
A key structural property is that node IDs in $D'$ are assigned in
insertion order; consequently, the relative ordering of any two non-variable
nodes in $D'$---and hence the operand semantics for non-commutative binary
operations---is fully determined by the sequence in which D2S inserts them.

\begin{definition}[Spiral Displacement Set]
\label{def:pairs}
For an output DAG of current size $n$, the candidate displacement set is
\[
  \mathcal{P}_n = \bigl\{(a,b)\in\mathbb{Z}^2 : -n \le a,b \le n\bigr\},
\]
sorted by the key $\bigl(|a|+|b|,\;|a|,\;(a,b)\bigr)$ (total cost, then
lexicographic tie-breaking).
This \emph{spiral enumeration} guarantees that lower-cost displacements are
tried first.
\end{definition}

The greedy loop iterates over $\mathcal{P}_n$ and, for each pair $(a,b)$,
evaluates four candidate operations in strict priority order:
$\texttt{V}\ell$ (insert via primary), $\texttt{v}\ell$ (insert via secondary), $\texttt{C}$, $\texttt{c}$.
The first valid operation is emitted, pointers are updated, and the loop
continues until all nodes and edges of $D$ are placed.

\begin{figure}[hbtp]
    \centering
    \includegraphics[width=\linewidth]{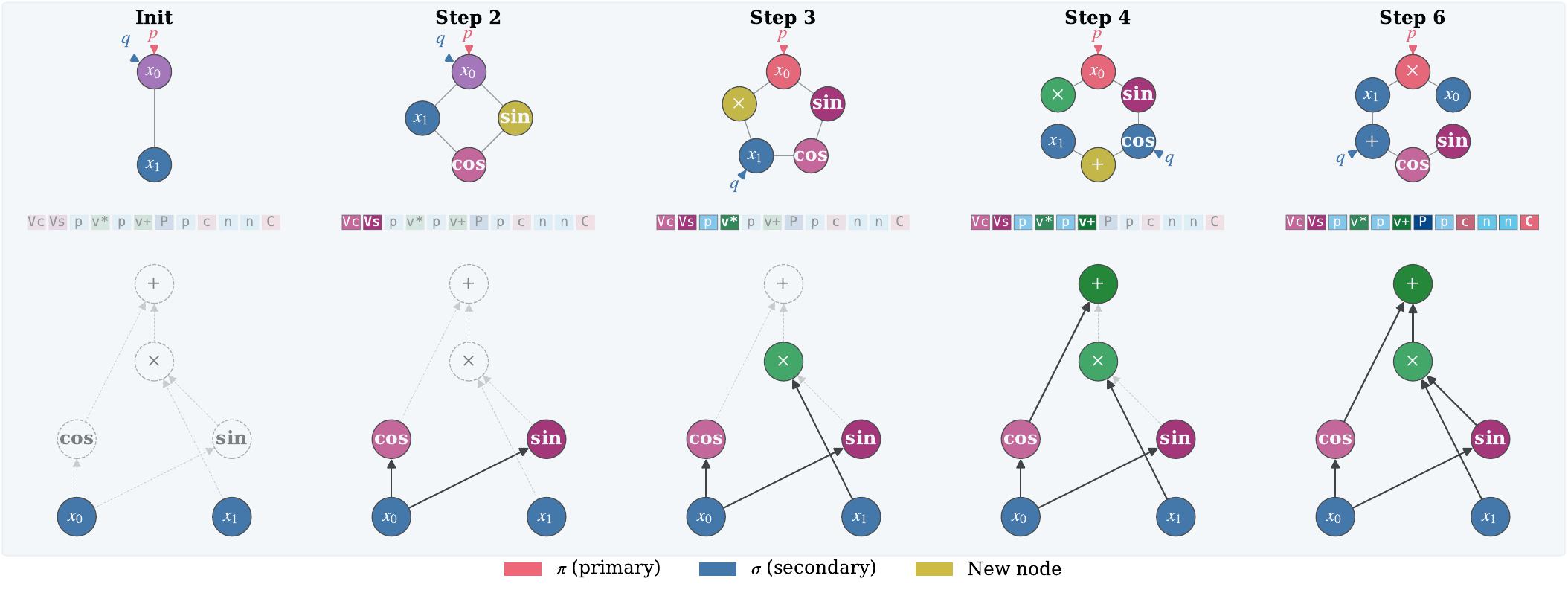}
    \caption{DAG-to-String encoding of $\sin(x_0) \cdot x_1 + \cos(x_0)$ via greedy traversal from
  $x_0$. Ghost nodes (dashed) indicate parts not yet encoded; the emitted token sequence
  converges to the canonical string as all nodes and edges are visited.}
    \label{fig:d2s}
\end{figure}

\begin{table}[ht]
\centering
\caption{Pseudocode for the DAG-to-String (D2S) algorithm.
  $\mathtt{moves}(a)$: emits $|a|$ copies of $\texttt{N}$ (if $a{>}0$) or $\texttt{P}$
  (if $a{<}0$).
  $\mathtt{moves}_s(b)$: analogously $\texttt{n}$/$\texttt{p}$ for the secondary pointer.}
\label{tab:d2s_pseudo}
\begin{tabular}{l}
\toprule
\textbf{Algorithm D2S}$(D,\,m)$ \\
\midrule
\textbf{Input:} labeled DAG $D=(V,E,\ell,\delta)$;\; number of variables $m$ \\
\textbf{Output:} IsalSR instruction string $w$ \\[3pt]
$w \leftarrow \varepsilon$;\;
Create empty output DAG $D'$;\;
$\mathtt{i2o} \leftarrow \{\}$;\; $\mathtt{o2i} \leftarrow \{\}$ \\
Map all $m$ \textsc{Var} nodes from $D$ to $D'$ in variable-index order;
  insert into CDLL $L$ \\
$p \leftarrow \mathtt{ptr}(x_1)$;\; $q \leftarrow \mathtt{ptr}(x_1)$ \\
$N_{\mathrm{left}} \leftarrow |V|-m$;\; $E_{\mathrm{left}} \leftarrow |E|$ \\[4pt]
\textbf{while} $N_{\mathrm{left}}>0$ \textbf{or} $E_{\mathrm{left}}>0$ \textbf{do} \\
\quad \textbf{foreach} $(a,b)\in\mathcal{P}_{|V'|}$ \textbf{do}
  \hfill\textit{// sorted by cost} \\
\qquad $\tilde{p} \leftarrow \mathtt{walk}(L,p,a)$;\;
       $u_p \leftarrow \mathtt{o2i}[\mathtt{val}(L,\tilde{p})]$ \\
\qquad \textbf{if} $N_{\mathrm{left}}>0$
  \textbf{and} $\exists$ uninserted $v\in N^+(u_p)$ \textbf{then} \\
\qquad\quad $w \mathrel{+}= \mathtt{moves}(a)\,\texttt{V}\ell_v$;\;
            Insert $v$ into $D'$ and CDLL after $\tilde{p}$;\;
            $p \leftarrow \tilde{p}$ \\
\qquad\quad Update $\mathtt{i2o},\mathtt{o2i}$;\;
            $N_{\mathrm{left}}{-}{-}$;\; $E_{\mathrm{left}}{-}{-}$;\;
            \textbf{break} \\
\qquad $\tilde{q} \leftarrow \mathtt{walk}(L,q,b)$;\;
       $u_q \leftarrow \mathtt{o2i}[\mathtt{val}(L,\tilde{q})]$ \\
\qquad \textbf{if} $N_{\mathrm{left}}>0$
  \textbf{and} $\exists$ uninserted $v\in N^+(u_q)$ \textbf{then} \\
\qquad\quad $w \mathrel{+}= \mathtt{moves}_s(b)\,\texttt{v}\ell_v$;\;
            Insert $v$ into $D'$ and CDLL after $\tilde{q}$;\;
            $q \leftarrow \tilde{q}$ \\
\qquad\quad Update $\mathtt{i2o},\mathtt{o2i}$;\;
            $N_{\mathrm{left}}{-}{-}$;\; $E_{\mathrm{left}}{-}{-}$;\;
            \textbf{break} \\
\qquad \textbf{if} $(u_p,u_q)\in E$
  \textbf{and} $(\mathtt{i2o}[u_p],\mathtt{i2o}[u_q])\notin E'$ \textbf{then} \\
\qquad\quad $w \mathrel{+}= \mathtt{moves}(a)\,\mathtt{moves}_s(b)\,\texttt{C}$;\;
            $E' \leftarrow E'\cup\{(\mathtt{i2o}[u_p],\mathtt{i2o}[u_q])\}$ \\
\qquad\quad $p \leftarrow \tilde{p}$;\; $q \leftarrow \tilde{q}$;\;
            $E_{\mathrm{left}}{-}{-}$;\; \textbf{break} \\
\qquad \textbf{if} $(u_q,u_p)\in E$
  \textbf{and} $(\mathtt{i2o}[u_q],\mathtt{i2o}[u_p])\notin E'$ \textbf{then} \\
\qquad\quad $w \mathrel{+}= \mathtt{moves}(a)\,\mathtt{moves}_s(b)\,\texttt{c}$;\;
            $E' \leftarrow E'\cup\{(\mathtt{i2o}[u_q],\mathtt{i2o}[u_p])\}$ \\
\qquad\quad $p \leftarrow \tilde{p}$;\; $q \leftarrow \tilde{q}$;\;
            $E_{\mathrm{left}}{-}{-}$;\; \textbf{break} \\[4pt]
\textbf{return} $w$ \\
\bottomrule
\end{tabular}
\end{table}

\subsection{Pruned Canonical String}
\label{sec:canonical}

The greedy D2S string depends on the internal node-numbering of the input
DAG; two structurally equivalent DAGs with different numberings may yield
different strings, and---crucially---different numberings may reverse the
operand order of non-commutative binary operations.
The \emph{canonical string} eliminates this dependency by exploring
\emph{all valid insertion orderings} at each $\texttt{V}\ell$/$\texttt{v}\ell$ branch point,
thereby also exploring all possible operand orders for non-commutative nodes.

\begin{definition}[Valid String Set]
\label{def:valid_set}
Let $\mathcal{W}(D)$ denote the set of all strings producible by the D2S
procedure on $D$ when, at each $\texttt{V}\ell$/$\texttt{v}\ell$ branch point with $k\geq 2$
uninserted out-neighbors of the tentative pointer position, any one of those
neighbors may be chosen as the next inserted node.
\end{definition}

\begin{definition}[Canonical String]
\label{def:canonical}
The \emph{canonical string} of labeled DAG $D$ is:
\[
  w^*_D \;=\; \mathrm{lexmin}\Bigl\{w \in \arg\min_{w'\in\mathcal{W}(D)}|w'|\Bigr\},
\]
i.e., the lexicographically smallest element among all shortest strings in
$\mathcal{W}(D)$, under a fixed total order on $\Sigma_{\mathrm{SR}}$.
\end{definition}

\noindent\textbf{6-Component structural tuple.}
To prune the exponential backtracking search while preserving the invariant
property, we associate with each node a directed structural descriptor.

\begin{definition}[6-Component Structural Tuple]
\label{def:6tuple}
For node $v$ in labeled DAG $D$, define the directed $k$-hop neighborhoods:
\begin{align*}
  \mathrm{out}_{N_k}(v) &= \{u\in V: \text{shortest directed path }
    v\leadsto u \text{ has length exactly } k\}, \\
  \mathrm{in}_{N_k}(v)  &= \{u\in V: \text{shortest directed path }
    u\leadsto v \text{ has length exactly } k\}.
\end{align*}
The \emph{6-component structural tuple} of $v$ is:
\[
  \tau(v) = \bigl(|\mathrm{in}_{N_1}(v)|,\;|\mathrm{out}_{N_1}(v)|,\;
                  |\mathrm{in}_{N_2}(v)|,\;|\mathrm{out}_{N_2}(v)|,\;
                  |\mathrm{in}_{N_3}(v)|,\;|\mathrm{out}_{N_3}(v)|\bigr)
  \in \mathbb{N}^6.
\]
All components are computed by truncated BFS in each direction up to depth\,3,
at cost $O(|V|+|E|)$ per node.
\end{definition}

\begin{definition}[Pruned Canonical String]
\label{def:pruned_canonical}
The \emph{pruned canonical string} $w^{**}_D$ is computed by the same
backtracking search as $w^*_D$, with two additional restrictions at each
$\texttt{V}\ell$/$\texttt{v}\ell$ branch point.

\textbf{Restriction 1 (first-operand eligibility).}
When the acting pointer node is $u$ and the candidate is a \textsc{Pow}
node $c$ that already has at least one in-neighbor recorded in $\sigma(c)$,
$c$ is eligible for $\texttt{V}\ell$/$\texttt{v}\ell$ insertion via $u$ only if $u$ would
become $\sigma(c)[0]$---the base.
Formally, $c$ is excluded if $\sigma(c)$ is non-empty and
$\sigma(c)[0]\neq u$.
This ensures that each \textsc{Pow} node is created via its designated
base source.

\textbf{Restriction 2 (label-aware $\tau$-pruning).}
Given the eligible candidate set $\mathcal{C}$ after applying Restriction~1,
candidates are partitioned by label:
$\mathcal{C}_{\ell_0} = \{c\in\mathcal{C}: \ell(c)=\ell_0\}$
for each distinct label $\ell_0$.
Within each group, only the maximum-tuple candidates are retained:
\[
  \mathcal{C}^*_{\ell_0}
  = \bigl\{c\in\mathcal{C}_{\ell_0}:
      \tau(c) = \max_{c'\in\mathcal{C}_{\ell_0}}\tau(c')\bigr\}.
\]
The explored candidate set is $\mathcal{C}^* = \bigcup_{\ell_0} \mathcal{C}^*_{\ell_0}$.
The selection criterion (shortest string, then lexmin) is applied over all
branches in $\mathcal{C}^*$.
\end{definition}

\noindent The complete algorithm is given in Table~\ref{tab:canon_pseudo},
using in-place state mutation with explicit undo on backtrack.

\begin{table}[ht]
\centering
\caption{Pseudocode for the Pruned Canonical String algorithm.
  \textsc{PCSstep} is called recursively and uses in-place mutation of
  $D'$, $L$, and the index maps, with explicit undo on each backtrack step.
  $\mathtt{eligible}(u, c)$ holds iff $c$ is not a binary non-commutative
  node with a pre-existing first operand other than $u$.}
\label{tab:canon_pseudo}
\begin{tabular}{l}
\toprule
\textbf{Algorithm PrunedCanonical}$(D,\,m)$ \\
\midrule
\textbf{Input:} labeled DAG $D=(V,E,\ell,\delta,\sigma)$;\; number of variables $m$ \\
\textbf{Output:} pruned canonical string $w^{**}_D$ \\[3pt]
$D \leftarrow \mathtt{normalize\_const\_creation}(D)$
  \hfill\textit{// redirect all \textsc{Const} creation edges to $x_1$} \\
Precompute $\tau(v)$ for all $v\in V$
  \hfill\textit{// $O(|V|(|V|+|E|))$} \\
Initialize $D'$, $L$, $\mathtt{i2o}$, $\mathtt{o2i}$, $p$, $q$
  as in D2S \\
\textbf{return}
  $\textsc{PCSstep}(D,D',L,p,q,\mathtt{i2o},\mathtt{o2i},|V|{-}m,|E|,\varepsilon,\tau)$ \\[4pt]
\midrule
\textbf{function}
  $\textsc{PCSstep}(D,D',L,p,q,\mathtt{i2o},\mathtt{o2i},
  N_{\mathrm{left}},E_{\mathrm{left}},\mathit{pfx},\tau)$: \\
\quad \textbf{if} $N_{\mathrm{left}}{=}0$ \textbf{and}
               $E_{\mathrm{left}}{=}0$ \textbf{then return} $\mathit{pfx}$ \\
\quad $\mathit{best} \leftarrow \bot$ \\
\quad \textbf{foreach} $(a,b)\in\mathcal{P}_{|V'|}$ \textbf{do} \\
\qquad $\tilde{p} \leftarrow \mathtt{walk}(L,p,a)$;\;
       $u_p \leftarrow \mathtt{o2i}[\mathtt{val}(L,\tilde{p})]$ \\
\qquad \textbf{if} $N_{\mathrm{left}}>0$ \textbf{then} \\
\qquad\quad $\mathcal{C} \leftarrow \{v:(u_p,v)\in E,\,v\notin\mathtt{i2o},\,
  \mathtt{eligible}(u_p,v)\}$
  \hfill\textit{// B9: first-operand filter} \\
\qquad\quad \textbf{if} $\mathcal{C} \neq \emptyset$ \textbf{then} \\
\qquad\qquad \textit{// label-aware pruning} \\
\qquad\qquad \textbf{foreach} label $\ell_0$ in $\{\ell(c):c\in\mathcal{C}\}$ \textbf{do} \\
\qquad\qquad\quad $G \leftarrow \{c\in\mathcal{C}:\ell(c)=\ell_0\}$ \\
\qquad\qquad\quad $\mathcal{C}^*_{\ell_0} \leftarrow
  \{c\in G:\tau(c)=\max_{c'\in G}\tau(c')\}$ \\
\qquad\qquad $\mathcal{C}^* \leftarrow \bigcup_{\ell_0}\mathcal{C}^*_{\ell_0}$ \\
\qquad\qquad \textbf{foreach} $c \in \mathcal{C}^*$ \textbf{do} \\
\qquad\qquad\quad \textit{Forward:} add $c$ to $D'$ and CDLL; update maps \\
\qquad\qquad\quad $r \leftarrow \textsc{PCSstep}(\ldots,\,
  \mathit{pfx}+\mathtt{moves}(a)+\texttt{V}\ell_c,\,\ldots,\,
  N_{\mathrm{left}}{-}1,\,E_{\mathrm{left}}{-}1,\,\ldots)$ \\
\qquad\qquad\quad \textbf{if} $\mathit{best}{=}\bot$
  \textbf{or} $(|r|,r)<(|\mathit{best}|,\mathit{best})$
  \textbf{then} $\mathit{best} \leftarrow r$ \\
\qquad\qquad\quad \textit{Undo:} remove $c$ from $D'$, CDLL, and index maps \\
\qquad\quad \textbf{return} $\mathit{best}$ \\
\qquad \textit{(symmetric block for $\texttt{v}\ell$ using $q$, displacement $b$,
               and the same label-aware filter)} \\
\qquad \textit{(deterministic $\texttt{C}$/$\texttt{c}$ blocks, same as D2S; no branching)} \\
\bottomrule
\end{tabular}
\end{table}

\subsection{Canonical String as a Graph Invariant}
\label{sec:invariant}

We now prove the central theoretical result of \textsc{IsalSR}.
The key step is a definition of labeled-DAG isomorphism that faithfully
captures the semantic equivalence of two expression DAGs, including the
operand-order convention for non-commutative operations.

\begin{definition}[Labeled-DAG Isomorphism]
\label{def:isomorphism}
Two labeled DAGs $D_1=(V_1,E_1,\ell_1,\delta_1)$ and
$D_2=(V_2,E_2,\ell_2,\delta_2)$ are \emph{isomorphic}, written $D_1\cong D_2$,
if there exists a bijection $\phi:V_1\to V_2$ satisfying:
\begin{enumerate}[nosep, label=(\roman*)]
  \item \textbf{Edge preservation:}
    $(u,v)\in E_1 \Leftrightarrow (\phi(u),\phi(v))\in E_2$;
  \item \textbf{Label preservation:}
    $\ell_1(u) = \ell_2(\phi(u))$ for all $u\in V_1$;
  \item \textbf{Variable anchoring:}
    for every \textsc{Var} node $u\in V_1$,
    $\delta_1(u) = \delta_2(\phi(u))$,
    i.e., $\phi(x_i^{(1)}) = x_i^{(2)}$ for all $i=1,\ldots,m$;
  \item \textbf{Operand-order preservation:}
    for every \textsc{Pow} node $v\in V_1$
    with ordered input list $\sigma_1(v)=(u_1,u_2)$,
    \[
      \sigma_2(\phi(v)) = (\phi(u_1),\,\phi(u_2)).
    \]
    That is, $\phi$ maps the base of $v$ in $D_1$ to the base of $\phi(v)$
    in $D_2$, and the exponent to the exponent.
    All other operation nodes are commutative and impose no ordering
    constraint.
\end{enumerate}
\end{definition}

Two main results follow. They are stated as conjectures. Their proof is left as future work.

\begin{conjecture}[Round-Trip Fidelity]
\label{thm:roundtrip}
Let $w\in\Sigma_{\mathrm{SR}}^*$ with $m\geq 1$ variables and
$D=\mathrm{S2D}(w,m)$.
If every non-variable node of $D$ is reachable from some variable via
directed paths, then
\[
  D \;\cong\; \mathrm{S2D}\bigl(\mathrm{D2S}(D,\,x_1),\;m\bigr).
\]
\end{conjecture}

\begin{conjecture}[Pruned Canonical String is a Complete Labeled-DAG Invariant]
\label{thm:invariant}
Let $D_1$ and $D_2$ be labeled DAGs with $m\geq 1$ variables, both
satisfying the reachability condition of Conjecture~\ref{thm:roundtrip}.
Then:
\[
  w^{**}_{D_1} = w^{**}_{D_2}
  \quad\Longleftrightarrow\quad
  D_1 \cong D_2.
\]
\end{conjecture}

%% file: mario/computational_experiments.tex
\label{sec:benchmarks}

Conjectures~\ref{thm:roundtrip} and~\ref{thm:invariant} assert that the pruned canonical string is a complete labeled-DAG invariant, but their proofs are left as future work.
We therefore validate the representation empirically by testing five fundamental properties of the canonical string:
\begin{description}[nosep, leftmargin=1.2em, labelwidth=1em]
  \item[P1] \textbf{Round-trip fidelity.}
    Encoding a DAG to a string and decoding it back recovers an isomorphic DAG:
    $\mathrm{S2D}\bigl(\mathrm{D2S}(D),\,m\bigr) \cong D$
    (Conjecture~\ref{thm:roundtrip}).
  \item[P2] \textbf{DAG acyclicity.}
    Every DAG produced by $\mathrm{S2D}$ is acyclic (\S\ref{sec:s2d}).
  \item[P3] \textbf{Canonical invariance.}
    The pruned canonical string is a complete isomorphism invariant:
    $w^{**}_{D_1} = w^{**}_{D_2} \Leftrightarrow D_1 \cong D_2$
    (Conjecture~\ref{thm:invariant}).
    As a corollary, canonicalization is idempotent.
  \item[P4] \textbf{Evaluation preservation.}
    A round-tripped DAG evaluates identically to the original at every test point,
    up to a tolerance of $10^{-8}$.
  \item[P5] \textbf{Search space reduction.}
    Canonicalization collapses the $\Theta(k!)$ equivalent node-numbering representations
    of each expression into a single canonical form (\S\ref{sec:invariant}).
\end{description}
Properties P1--P4 are validated by a single-pass randomized experiment (\S\ref{sec:exp_properties}).
Property P5 is validated by a separate search-space analysis (\S\ref{sec:exp_search_space}).
A scalability study (\S\ref{sec:exp_timing}) measures the CPU time of the pruned canonicalization algorithm as a function of DAG size.
Two illustrative experiments (\S\ref{sec:exp_shortest_path}--\ref{sec:exp_neighborhood}) provide qualitative evidence that the canonical string induces a meaningful metric on expression DAGs.
All experiments use a global random seed of $42$ and are fully reproducible from the public code repository.
Results are presented in Section~\ref{sec:results}.

\subsection{Benchmark Suites}
\label{sec:benchmark_suites}

We draw benchmark expressions from two standard SR suites: the Nguyen benchmarks and a subset of the AI~Feynman equations.

\paragraph{Nguyen benchmarks.}
Table~\ref{tab:nguyen} lists the 12 Nguyen benchmark expressions originally introduced by \citet{uy2011} and adopted in the configuration of \citet{liu2025}.
Eight are single-variable polynomials, trigonometric, logarithmic, or radical functions over bounded intervals; four are two-variable expressions involving products, sums, and the sole non-commutative operator \textsc{Pow}.
Training data consist of $20$ points sampled uniformly from the specified input range; test data consist of $100$ points from the same range.

\begin{table}[htbp]
\centering
\caption{Nguyen symbolic regression benchmarks, following the configuration of \citet{liu2025}.
  Training: $20$ uniform samples; testing: $100$ uniform samples; seed $42$.}
\label{tab:nguyen}
\begin{tabular}{llcl}
\toprule
Name & Expression $f(x)$ or $f(x,y)$ & $m$ & Input range \\
\midrule
Nguyen-1  & $x^3 + x^2 + x$              & 1 & $[-1, 1]$ \\
Nguyen-2  & $x^4 + x^3 + x^2 + x$        & 1 & $[-1, 1]$ \\
Nguyen-3  & $x^5 + x^4 + x^3 + x^2 + x$  & 1 & $[-1, 1]$ \\
Nguyen-4  & $x^6 + x^5 + x^4 + x^3 + x^2 + x$ & 1 & $[-1, 1]$ \\
Nguyen-5  & $\sin(x^2)\cos(x) - 1$        & 1 & $[-1, 1]$ \\
Nguyen-6  & $\sin(x) + \sin(x + x^2)$     & 1 & $[-1, 1]$ \\
Nguyen-7  & $\ln(x{+}1) + \ln(x^2{+}1)$   & 1 & $[0, 2]$  \\
Nguyen-8  & $\sqrt{x}$                     & 1 & $[0, 4]$  \\
Nguyen-9  & $\sin(x) + \sin(y^2)$          & 2 & $[-1, 1]$ \\
Nguyen-10 & $2\sin(x)\cos(y)$              & 2 & $[-1, 1]$ \\
Nguyen-11 & $x^y$                          & 2 & $[0, 1]$  \\
Nguyen-12 & $x^4 - x^3 + y^2/2 - y$       & 2 & $[-1, 1]$ \\
\bottomrule
\end{tabular}
\end{table}

\paragraph{AI~Feynman benchmarks.}
Table~\ref{tab:feynman} lists 10 physics equations selected from the AI~Feynman dataset of \citet{udrescu2020}, following the subset and variable ranges of \citet{liu2025}.
The selection spans one single-variable, five two-variable, and four three-variable equations, covering polynomial, trigonometric, and rational functional forms.
Data consist of $200$ samples with an $80$/$20$ train/test split.

\begin{table}[htbp]
\centering
\caption{AI~Feynman benchmark subset, following the selection and variable ranges of \citet{liu2025}.
  Data: $200$ samples, $80$/$20$ train/test split, seed $42$.}
\label{tab:feynman}
\begin{tabular}{llcl}
\toprule
ID & Expression & $m$ & Variable ranges \\
\midrule
I.6.20a  & $e^{-\theta^2/2}/\!\sqrt{2\pi}$  & 1 & $\theta \in [1, 3]$ \\
I.12.1   & $\mu \, N_s$                       & 2 & $[1, 5]^2$ \\
I.25.13  & $q / C$                            & 2 & $[1, 3]^2$ \\
I.34.27  & $\hbar\,\omega$                    & 2 & $[1, 5]^2$ \\
I.39.10  & $p_r V / 2$                        & 2 & $[1, 5]^2$ \\
II.3.24  & $p\,r / (4\pi)$                    & 2 & $[1, 5]^2$ \\
I.14.3   & $m\,g\,z$                          & 3 & $[1, 5]^3$ \\
I.12.4   & $q_1 / (4\pi r c)$                 & 3 & $[1, 5]^3$ \\
I.10.7   & $m_0 / \!\sqrt{1 - v^2/c^2}$       & 3 & $m_0 {\in} [1,5],\; v {\in} [1,2],\; c {\in} [3,10]$ \\
I.48.20  & $mc^2 / \!\sqrt{1 - (v/c)^2}$      & 3 & $m {\in} [1,5],\; c {\in} [1,2],\; v {\in} [3,10]$ \\
\bottomrule
\end{tabular}
\end{table}

\subsection{Random Expression Generation}
\label{sec:random_generation}

Two complementary procedures generate test inputs across the experiments.

\paragraph{Random DAG generation.}
The scalability (\S\ref{sec:exp_timing}) and search-space (\S\ref{sec:exp_search_space}) experiments require random DAGs with a prescribed number of internal nodes.
Given $m$ input variables and a target of $k$ internal nodes, the generator creates $m$ \textsc{Var} leaf nodes and then iteratively adds $k$ operation nodes.
At each step, with probability $0.6$ a unary operation is selected uniformly from $\{\textsc{Sin}, \textsc{Cos}, \textsc{Exp}, \textsc{Log}, \textsc{Abs}\}$; otherwise a binary operation is selected from $\{\textsc{Add}, \textsc{Mul}, \textsc{Sub}, \textsc{Div}\}$.
If fewer than two nodes exist, a unary operation is forced.
Each new node receives one (unary) or two (binary) incoming edges from nodes selected uniformly at random among all previously created nodes.
\textsc{Pow} is excluded by default, so all binary operations are commutative under the decomposition of \S\ref{sec:iset} ($\textsc{Sub} \to \textsc{Add} + \textsc{Neg}$, $\textsc{Div} \to \textsc{Mul} + \textsc{Inv}$).
The per-sample seed is deterministic: $\mathrm{seed} = 42 + m \cdot 10{,}000 + k \cdot 100 + s$, where $s$ is the sample index.

\paragraph{Random string generation.}
The property-validation experiment (\S\ref{sec:exp_properties}) generates random IsalSR strings as its input.
A random string is produced by first sampling a token count $n \sim \mathrm{Uniform}(1, T_{\max})$ and then drawing each token independently and uniformly from the full IsalSR token set (Definition~\ref{def:alphabet}).
Each string is parsed via $\mathrm{S2D}$ (Table~\ref{tab:s2d_pseudo}); strings that produce DAGs with only \textsc{Var} nodes (no internal nodes) are discarded.
This strategy samples the string space uniformly rather than targeting a particular DAG topology, exercising all token types including movement, edge creation, and no-op instructions.

\subsection{Property Validation: P1--P4}
\label{sec:exp_properties}

We validate properties P1--P4 in a single pass over randomly generated IsalSR strings.
For each $m \in \{1, 2, 3\}$ input variables, we generate $N = 5{,}000$ random strings with $T_{\max} = 20$ tokens.
Each valid, non-VAR-only DAG $D$ undergoes the following four tests:
\begin{enumerate}[nosep]
  \item \textbf{P2 (Acyclicity):} A topological sort of $D$ is computed; the test passes if the sort includes all $|V|$ nodes.
  \item \textbf{P1 (Round-trip):} The greedy encoding $w' = \mathrm{D2S}(D)$ is decoded to $D' = \mathrm{S2D}(w', m)$; the test passes if $D \cong D'$ under Definition~\ref{def:isomorphism}.
  \item \textbf{P4 (Evaluation preservation):} Both $D$ and the round-tripped $D'$ are evaluated at five fixed test points per $m$ configuration, spanning negative, zero, and positive inputs.
    The test passes if the maximum absolute error across all test points is below $10^{-8}$.
    When both DAGs produce the same evaluation error at a test point (e.g., division by zero), that point is counted as preserved.
  \item \textbf{P3 (Canonical invariance):} The pruned canonical string $w^{**} = \mathrm{canonical}(D)$ is decoded to $D'' = \mathrm{S2D}(w^{**}, m)$; invariance passes if $D \cong D''$.
    Canonicalization is then applied a second time to $D''$; idempotence passes if the resulting string equals $w^{**}$.
    The canonicalization timeout is $2.0$\,s; samples that exceed the timeout are excluded from the P3 denominator but remain in P1, P2, and P4 counts.
\end{enumerate}
Pass rates are reported with Clopper--Pearson exact $95\%$ binomial confidence intervals, for a total of $3 \times 5{,}000 = 15{,}000$ random strings.

As a complementary check, properties P1--P4 are also verified deterministically on a subset of $8$ benchmark expressions drawn from the Nguyen (Table~\ref{tab:nguyen}) and AI~Feynman (Table~\ref{tab:feynman}) suites, spanning $m \in \{1, 2, 3\}$ input variables and DAG sizes from $|V| = 3$ to $|V| = 14$.

\subsection{Property Validation: P5 --- Search Space Reduction}
\label{sec:exp_search_space}

This experiment directly validates the central claim: that canonicalization collapses the $\Theta(k!)$ equivalent node-numbering representations of each expression into a single canonical form.

Rather than sampling random strings and measuring collision rates---an approach that underestimates the true equivalence class size because the string space is vastly larger than the number of structurally unique DAGs---we deliberately construct all equivalent representations of each expression and verify that canonicalization maps them to the same string.

A labeled DAG $D = (V, E, \ell, \delta)$ with $m$ variable nodes (IDs $0, \ldots, m{-}1$) and $k$ internal nodes (IDs $m, \ldots, m{+}k{-}1$) admits $k!$ distinct \emph{node-numbering schemes}: permutations $\pi$ of the internal node IDs that produce structurally distinct labeled DAGs, all encoding the same mathematical expression.
If the canonical string is a complete isomorphism invariant, it must map all $k!$ numbering schemes to the same string.

For each value of $k \in \{1, \ldots, 12\}$ and $m \in \{1, 2\}$ input variables, we generate structurally unique random DAGs (\S\ref{sec:random_generation}), deduplicating by canonical string to ensure that each DAG represents a distinct isomorphism class.
For each DAG $D$ with $k$ internal nodes, we proceed as follows:
\begin{enumerate}[nosep]
  \item Compute the pruned canonical string $w^{**} = \mathrm{canonical}(D)$ once.
  \item Generate permutations: all $k!$ permutations if $k \leq 8$ ($k! \leq 40{,}320$); otherwise $50{,}000$ random permutations.
  \item For each permutation $\pi$ of $\{0, 1, \ldots, k{-}1\}$, construct $D_\pi$ by remapping internal node $m{+}i$ to position $m{+}\pi(i)$, preserving all edges, labels, and operand order.
  Compute a \emph{structural fingerprint}: the tuple $\bigl(\ell(v),\, \sigma(v)\bigr)$ for each node $v$ in new-ID order, where $\sigma(v)$ is the ordered input list (Definition~\ref{def:ldag}).
  \item Verify \emph{canonical invariance} on a subset of $100$ permutations: compute $\mathrm{canonical}(D_\pi)$ and check that it equals $w^{**}$.
\end{enumerate}

Two distinct permutations $\pi_1, \pi_2$ produce the same structural fingerprint if and only if they differ by an automorphism of $D$.
By the Orbit-Stabilizer theorem, the number of distinct fingerprints is exactly
\begin{equation}
\label{eq:orbit_stabilizer}
  n_{\text{distinct}} = \frac{k!}{\lvert\mathrm{Aut}(D)\rvert},
\end{equation}
where $\lvert\mathrm{Aut}(D)\rvert$ is the size of the automorphism group of $D$.
For generic DAGs with no symmetry ($\lvert\mathrm{Aut}(D)\rvert = 1$), this equals $k!$.
The normalized ratio $n_{\text{distinct}} / k!$ therefore measures the structural asymmetry of each DAG, and we report it alongside the canonical invariance success rate.

\subsection{Canonicalization Scalability}
\label{sec:exp_timing}

This experiment measures how the wall-clock time of the pruned canonicalization algorithm (Definition~\ref{def:pruned_canonical}) scales with DAG size, and quantifies the speedup that $\tau$-tuple pruning provides over exhaustive search (Definition~\ref{def:canonical}).

We generate random DAGs in a full factorial design over $k \in \{1, 2, \ldots, 15\}$ internal nodes and $m \in \{1, 2, 3\}$ input variables, with $200$ independently seeded samples per $(k, m)$ configuration, for a total of $15 \times 3 \times 200 = 9{,}000$ DAGs.
Each DAG is canonicalized twice: once by the exhaustive backtracking algorithm and once by the pruned algorithm.
Both calls are subject to a per-canonicalization timeout of $120$\,s.
Wall-clock time is measured via a monotonic high-resolution counter.

Because canonicalization times are heavily right-skewed, we report medians and interquartile ranges (IQR) rather than means.
The \emph{speedup ratio} is defined as the median exhaustive time divided by the median pruned time for each $(k, m)$ configuration.

\subsection{Metric Space Properties}
\label{sec:exp_metric}

If the pruned canonical string is a complete invariant (Conjecture~\ref{thm:invariant}), the Levenshtein edit distance~\citep{levenshtein1966} on canonical strings satisfies the identity of indiscernibles and defines a metric on isomorphism classes of expression DAGs.
We probe this metric with two illustrative experiments.

\subsubsection{Shortest-Path Distance Between Expression DAGs}
\label{sec:exp_shortest_path}

We select five expression pairs that span a range of structural relationships:
(A)~structurally different expressions ($\sin(x)$ vs.\ $x^2 + x$),
(B)~identical operand structure with a different binary combinator ($\sin(x) + \cos(x)$ vs.\ $\sin(x) \cdot \cos(x)$),
(C)~a containment relationship ($x^2$ vs.\ $x^3 + x^2 + x$),
(D)~identical topology with a different unary operator ($\exp(x)$ vs.\ $\log(x)$), and
(E)~mixed two-variable expressions ($\sin(x) + y^2$ vs.\ $\cos(x) \cdot y$).
For each pair, both expressions are converted to labeled DAGs, their pruned canonical strings $w^{**}$ are computed, and the Levenshtein distance is obtained via the Wagner--Fischer dynamic-programming algorithm with full edit-operation backtrace.
This experiment is illustrative and involves no randomization or statistical analysis.

\subsubsection{Distance-1 Neighbourhood Analysis}
\label{sec:exp_neighborhood}

This experiment quantifies the redundancy that canonicalization eliminates in the immediate neighbourhood of a canonical string.
We take the expression $\sin(x_1) + \cos(x_1)$ with $m = 1$ variable, whose pruned canonical string is $w^{**} = \texttt{VcVspv{+}Ppc}$ of length $L = 10$.

We enumerate all character-level Levenshtein distance-$1$ neighbours of $w^{**}$.
Let $|\mathcal{A}|$ denote the character-level alphabet size (here $|\mathcal{A}| = 17$, comprising 7 single-character instruction tokens and 10 label characters that appear as the second character of compound tokens).
The total number of neighbours is:
\begin{equation}
\label{eq:neighborhood_size}
  N_{\text{total}} = \underbrace{L}_{\text{deletions}}
  + \underbrace{L \cdot (|\mathcal{A}| - 1)}_{\text{substitutions}}
  + \underbrace{(L + 1) \cdot |\mathcal{A}|}_{\text{insertions}}
  = 10 + 160 + 187 = 357.
\end{equation}
Each neighbour string is parsed via $\mathrm{S2D}$; strings that fail to parse or produce VAR-only DAGs are classified as invalid.
Valid DAGs are canonicalized (timeout $5$\,s), and the number of distinct canonical forms is counted.
The \emph{redundancy rate} is defined as $1 - N_{\text{unique}} / N_{\text{valid}}$, where $N_{\text{unique}}$ is the number of distinct canonical strings among the $N_{\text{valid}}$ valid neighbours.
Results are further disaggregated by edit operation type (deletion, substitution, insertion).
This experiment is a single-expression case study.

\subsection{Statistical Methods and Computational Infrastructure}
\label{sec:stats_infra}

\paragraph{Statistical methods.}
The property-validation experiment (\S\ref{sec:exp_properties}) reports Clopper--Pearson exact $95\%$ binomial confidence intervals for each pass rate, which are conservative and valid for all sample sizes.
The search-space experiment (\S\ref{sec:exp_search_space}) reports exact counts for exhaustive permutation runs ($k \leq 8$) and lower-bound counts for sampled runs ($k > 8$); the canonical invariance success rate is an exact binomial proportion.
The scalability experiment (\S\ref{sec:exp_timing}) reports medians and interquartile ranges, which are robust to the right-skewed distributions typical of execution-time measurements.
All experiments use a base random seed of $42$, with per-sample seeds derived deterministically from the base seed, the number of variables, the DAG size, and the sample index.

\paragraph{Computational infrastructure.}
The two illustrative experiments (\S\ref{sec:exp_shortest_path}--\ref{sec:exp_neighborhood}) are deterministic, single-expression computations that run in under one minute on a standard workstation.
The remaining experiments run on the Picasso supercomputer at the Supercomputing and Bioinnovation Center (SCBI) of the University of M\'alaga, using CPU-only nodes.
Table~\ref{tab:experiment_params} summarizes the parameters and computational scale of each experiment.

\begin{table}[htbp]
\centering
\caption{Summary of experiment parameters.
  $k$: internal nodes; $m$: input variables; $N$: samples per configuration.
  All experiments except the two illustrative ones run on the Picasso supercomputer.}
\label{tab:experiment_params}
\begin{tabular}{clcccrl}
\toprule
 & Experiment & $k$ & $m$ & $N$ & Timeout & Total \\
\midrule
P1--P4 & Property validation  & ---             & $1$--$3$  & 5{,}000/$m$        & 2\,s   & 15{,}000 \\
P5     & Search space         & $1$--$12$       & $1$--$2$  & ${\sim}40$/config  & 5\,s   & 961 \\
       & Scalability          & $1$--$15$       & $1$--$3$  & 200/config         & 120\,s & 9{,}000 \\
\midrule
       & Shortest path        & ---             & $1$--$2$  & 5 pairs            & ---    & 5 \\
       & Neighbourhood        & ---             & $1$       & 357 neighbours     & 5\,s   & 357 \\
\bottomrule
\end{tabular}
\end{table}

%% file: mario/results.tex
\begin{figure*}[t]
\centering
\includegraphics[width=\textwidth]{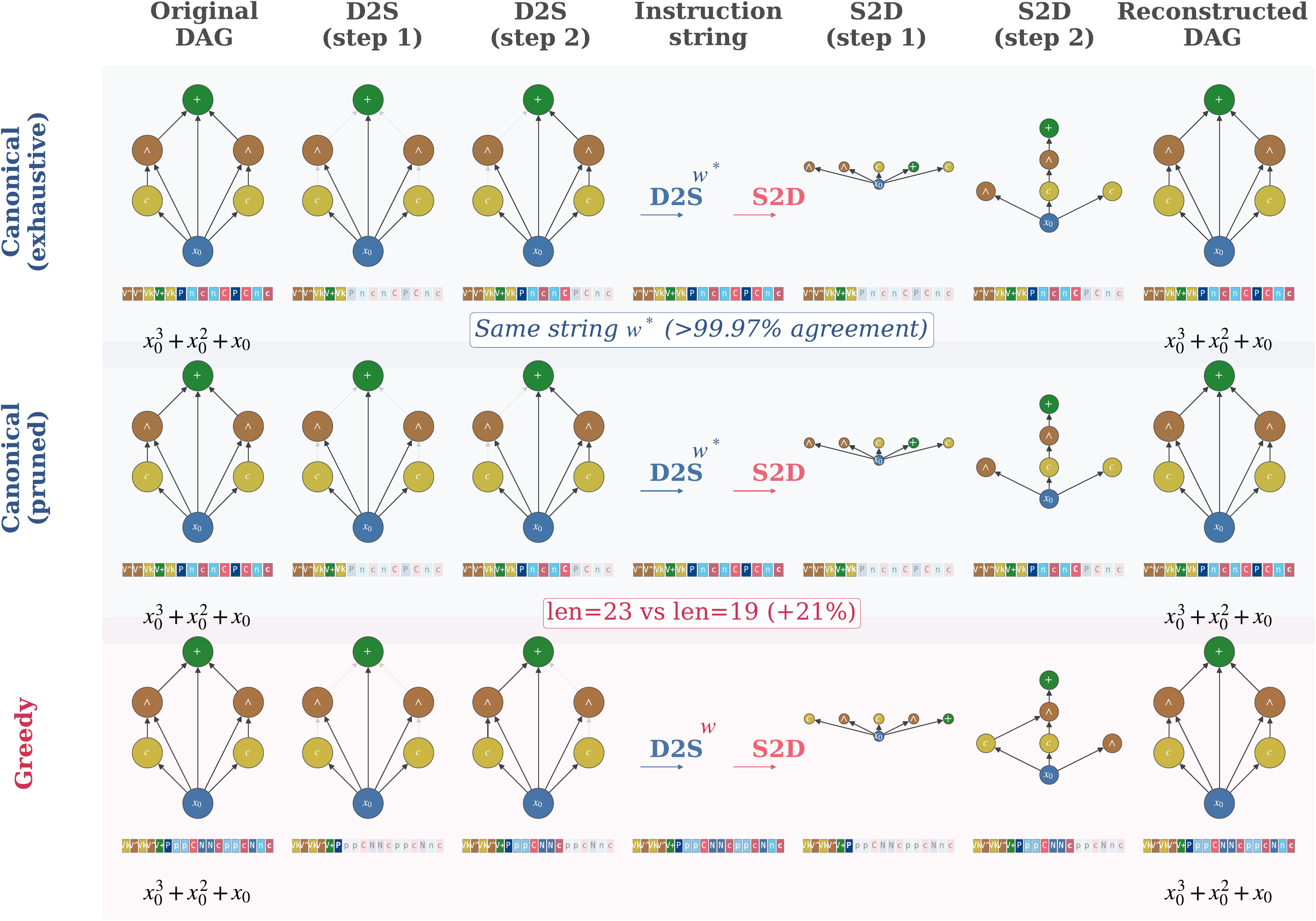}
\caption{Round-trip property for the Nguyen-1 benchmark ($x_0^3 + x_0^2 +
x_0$). Each row applies a different DAG-to-String (D2S) algorithm, followed by
String-to-DAG (S2D) reconstruction. \textbf{Columns~0,\,6}: original and
reconstructed DAGs (isomorphic, $\cong$). \textbf{Columns~1--2}: D2S
progressively encodes the DAG; ghost (dashed) nodes and edges indicate parts
not yet encoded, while the instruction string builds up left-to-right.
\textbf{Column~3}: the complete instruction string ($w^*$ for canonical, $w$
for greedy). \textbf{Columns~4--5}: S2D rebuilds the DAG from the string. The
canonical and pruned algorithms produce identical strings ($|w^*|{=}19$,
$>$99.97\% agreement), while the greedy algorithm yields a longer encoding
($|w|{=}23$, $+$21\%), yet all three round-trip to isomorphic DAGs.}
\label{fig:roundtrip}
\end{figure*}

\begin{figure*}[t]
\centering
\includegraphics[width=\textwidth]{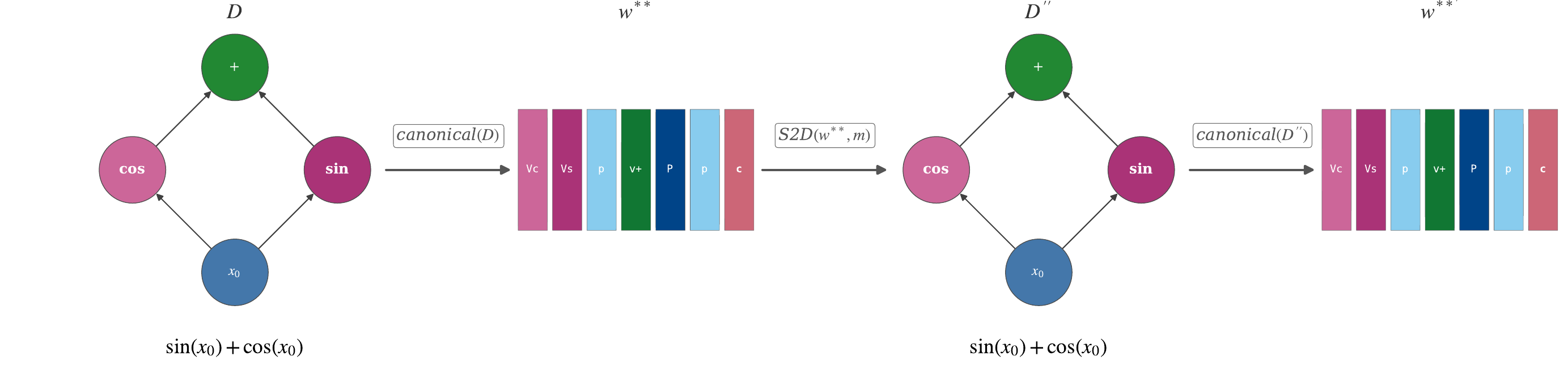}
\caption{Property~P3 (canonical invariance and idempotence) for $\sin(x_0) +
\cos(x_0)$. The original DAG~$D$ is canonicalized to produce $w^{**} =
\texttt{VcVspv+Ppc}$, then decoded via S2D to reconstruct~$D''$, which is
structurally isomorphic to~$D$ (\textbf{invariance}: $D \cong D''$). Applying
canonicalization a second time yields $w^{**'} = w^{**}$ (\textbf{idempotence}),
confirming that the canonical string is a fixed point of the
compose-and-recanonicalize map. Colour-coded token blocks show the instruction
string at each transformation step.}
\label{fig:p3_invariance}
\end{figure*}

\begin{figure*}[t]
\centering
\includegraphics[width=\textwidth]{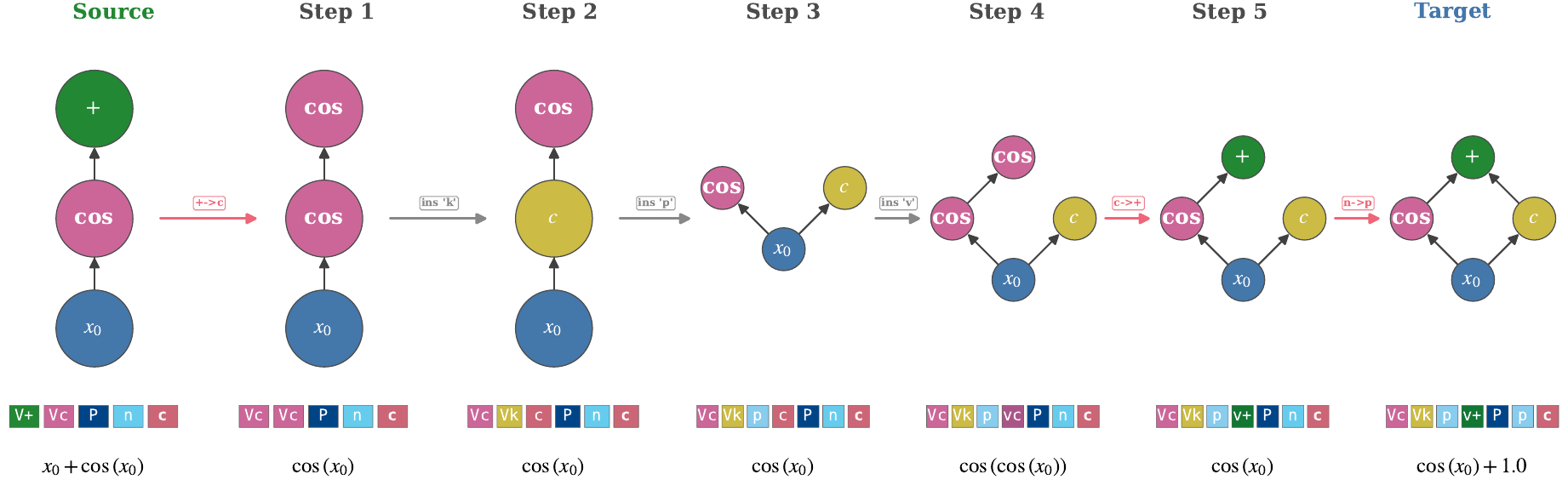}
\caption{Shortest Levenshtein path between $\cos(x_0) + x_0$
(\texttt{V+VcPnc}) and $\cos(x_0) + 1$ (\texttt{VcVkpv+Ppc}) in the canonical
string space ($d_{\mathrm{Lev}} = 6$). Each step applies one character-level
edit (substitution or insertion) to the current string. All intermediate strings
produce valid expression DAGs, progressing through simplified $\cos(x_0)$
forms---including a $\cos(\cos(x_0))$ detour at Step~4---before reaching the
target. Below each DAG: the corresponding IsalSR instruction string
(colour-coded by token type) and the mathematical expression.}
\label{fig:shortest_path}
\end{figure*}

\begin{figure*}[t]
\centering
\includegraphics[width=\textwidth]{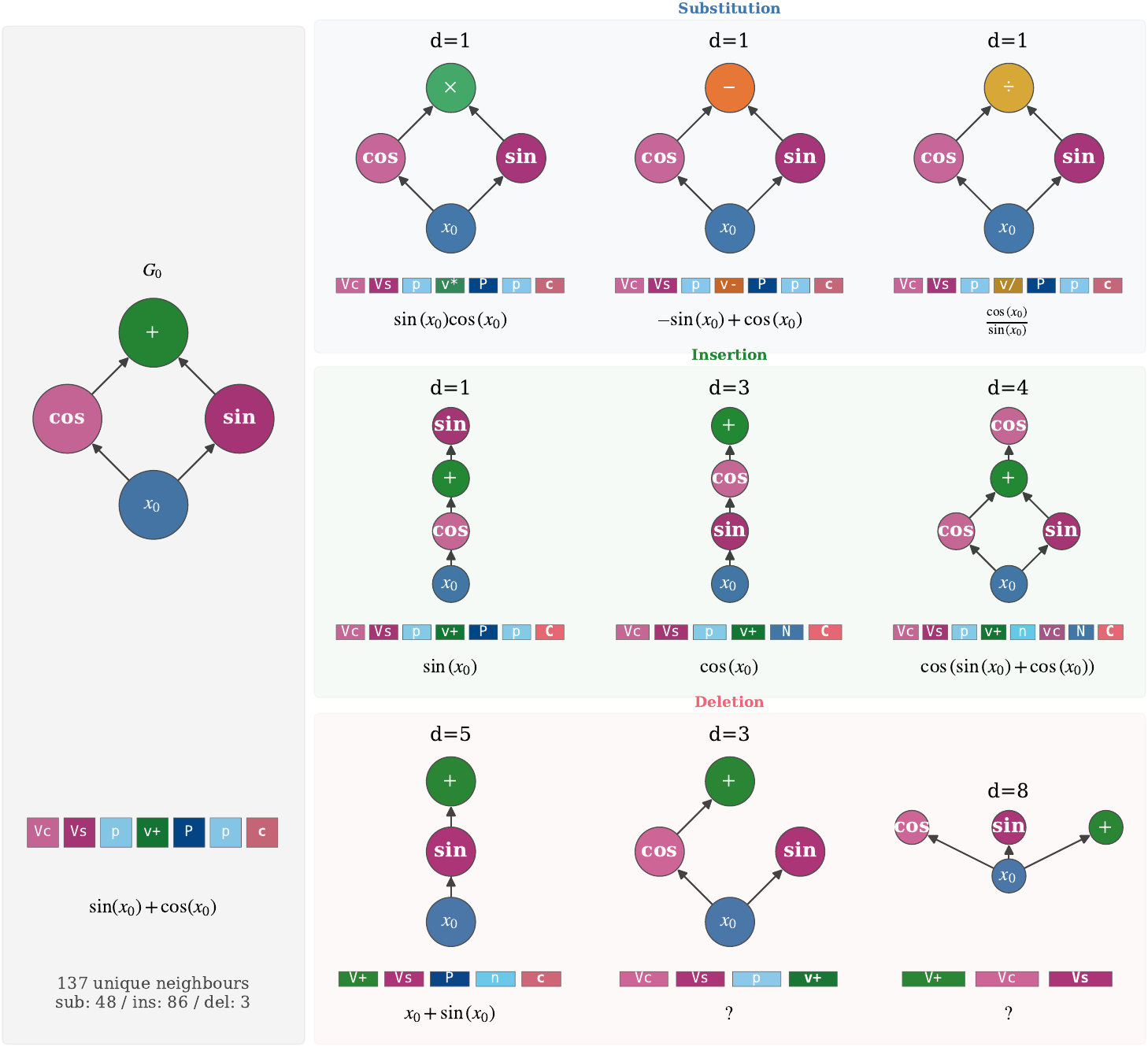}
\caption{Neighbourhood structure in the canonical string space around $G_0:
\sin(x_0) + \cos(x_0)$ (canonical string \texttt{VcVspv+Ppc}). Each neighbour is
obtained by a single Levenshtein edit (substitution, insertion, or deletion) on
the canonical string, followed by re-canonicalization. The $d$ labels indicate
the Levenshtein distance between the neighbour's canonical string and the base.
\textbf{Substitution} neighbours ($d{=}1$) change only the combinator operation
($+{\to}\times$, $+{\to}-$, $+{\to}\div$), preserving the operand structure.
\textbf{Insertion} and \textbf{deletion} neighbours produce structurally
different DAGs with varying distances. Of the 132 valid Lev-1 perturbations,
only 38 yield distinct canonical forms (71.2\% redundancy), illustrating the
$\Theta(k!)$ search-space reduction.}
\label{fig:neighbourhood}
\end{figure*}

We present the results of the experiments described in Section~\ref{sec:Computational-experiments}.
We first report the property-validation results (P1--P4) and the search-space reduction analysis (P5), then the scalability measurements, and finally the two metric-space case studies.

\subsection{Property Validation (P1--P4)}
\label{sec:results_properties}

We report the results of the randomized property-validation experiment described in \S\ref{sec:exp_properties}.
Of the $15{,}000$ random strings generated ($5{,}000$ per $m \in \{1,2,3\}$), $14{,}841$ produced valid DAGs with at least one internal node and constitute the test pool for properties P1, P2, and P4.
For P3, $5{,}713$ samples exceeded the $2.0$\,s canonicalization timeout and were excluded from the denominator, leaving $9{,}128$ evaluable samples.

\paragraph{P1 (Round-trip fidelity).}
All $14{,}841$ samples pass: $\mathrm{S2D}(\mathrm{D2S}(D),\,m) \cong D$ in every case.
The Clopper--Pearson $95\%$ confidence interval for the true pass rate is $[99.975\%,\;100\%]$.
Figure~\ref{fig:roundtrip} illustrates the round-trip for the Nguyen-1 benchmark, where the canonical and pruned algorithms produce identical strings ($|w^{**}| = 19$), while the greedy encoder yields a longer string ($|w| = 23$, a $21\%$ overhead).

\paragraph{P2 (DAG acyclicity).}
All $14{,}841$ decoded DAGs are acyclic, with the same confidence interval $[99.975\%,\;100\%]$.
P2 holds by construction: $\mathrm{S2D}$ assigns strictly increasing IDs to new nodes and rejects \texttt{C}/\texttt{c} edges that would create cycles (\S\ref{sec:s2d}).
The experiment confirms that the implementation matches this specification.

\paragraph{P4 (Evaluation preservation).}
All $14{,}841$ round-tripped DAGs evaluate identically to their originals at every test point (maximum absolute error below $10^{-8}$), with confidence interval $[99.975\%,\;100\%]$.
Cases where both original and round-tripped DAGs produce the same evaluation error (e.g., division by zero) are counted as preserved.

\paragraph{Benchmark validation.}
Table~\ref{tab:benchmark_properties} reports the deterministic validation of P1--P4 on $8$ benchmark expressions from the Nguyen and AI~Feynman suites.
All four properties hold for every expression, across DAG sizes ranging from $|V| = 3$ (Nguyen-8, $\sqrt{x}$) to $|V| = 14$ (Nguyen-12, a four-term polynomial in two variables).

\begin{table}[htbp]
\centering
\caption{Deterministic validation of P1--P4 on benchmark expressions.
  $|V|$: node count; $|E|$: edge count; $|w^{**}|$: pruned canonical string length.}
\label{tab:benchmark_properties}
\begin{tabular}{lcrrrccccc}
\toprule
Expression & $m$ & $|V|$ & $|E|$ & Depth & P1 & P2 & P3 & P4 & $|w^{**}|$ \\
\midrule
Nguyen-1    & 1 &  6 &  9 & 3 & \checkmark & \checkmark & \checkmark & \checkmark & 19 \\
Nguyen-5    & 1 &  8 & 10 & 5 & \checkmark & \checkmark & \checkmark & \checkmark & 26 \\
Nguyen-7    & 1 &  9 & 12 & 5 & \checkmark & \checkmark & \checkmark & \checkmark & 32 \\
Nguyen-8    & 1 &  3 &  3 & 2 & \checkmark & \checkmark & \checkmark & \checkmark &  7 \\
Nguyen-9    & 2 &  7 &  7 & 4 & \checkmark & \checkmark & \checkmark & \checkmark & 19 \\
Nguyen-10   & 2 &  6 &  6 & 2 & \checkmark & \checkmark & \checkmark & \checkmark & 16 \\
Nguyen-12   & 2 & 14 & 21 & 4 & \checkmark & \checkmark & \checkmark & \checkmark & 56 \\
Feynman-I.14.3 & 3 & 4 & 3 & 1 & \checkmark & \checkmark & \checkmark & \checkmark & 7 \\
\bottomrule
\end{tabular}
\end{table}

\paragraph{P3 (Canonical invariance).}
Of the $9{,}128$ non-timeout samples, $9{,}125$ pass ($99.97\%$).
Table~\ref{tab:p3_breakdown} reports the per-$m$ breakdown.

\begin{table}[htbp]
\centering
\caption{P3 pass rates by number of input variables $m$.
  Clopper--Pearson exact $95\%$ confidence intervals.}
\label{tab:p3_breakdown}
\begin{tabular}{crrcl}
\toprule
$m$ & Evaluated & Pass & Rate & $95\%$ CI \\
\midrule
1 & 3{,}004 & 3{,}004 & 100\%   & $[99.88\%,\;100\%]$ \\
2 & 3{,}062 & 3{,}061 & 99.97\% & $[99.82\%,\;100\%]$ \\
3 & 3{,}062 & 3{,}060 & 99.93\% & $[99.74\%,\;100\%]$ \\
\midrule
All & 9{,}128 & 9{,}125 & 99.97\% & $[99.89\%,\;100\%]$ \\
\bottomrule
\end{tabular}
\end{table}

Three samples fail P3: sample index $274$ at $k = 14$ for $m \in \{2, 3\}$ and sample index $2{,}603$ at $k = 6$ for $m = 3$.
None of these failures are timeouts.
We reproduced all three locally and identified a single root cause.

\paragraph{Root cause of P3 failures.}
In all three cases, the failing DAG contains an edge directed \emph{into} a \textsc{Var} node---a structure that can arise from $\mathrm{S2D}$ when a \texttt{C}/\texttt{c} token targets a variable position on the CDLL.
Concretely, in sample $2{,}603$ ($m = 3$, $k = 6$), a \textsc{Const} node (ID $6$) has an edge into \textsc{Var} node $x_0$ (ID $0$): the DAG contains the edge $(6, 0)$.
Similarly, sample $274$ contains an edge from a \textsc{Const} node into $x_0$ at $k = 14$.
These edges are semantically vacuous---\textsc{Var} nodes ignore their inputs during evaluation, which explains why P4 (evaluation preservation) passes---but they create a structure that the canonical $\mathrm{D2S}$ algorithm cannot encode.
The algorithm terminates with a \texttt{RuntimeError} rather than a timeout, because no candidate node satisfies the insertion conditions when the remaining unencoded edges include an edge into a leaf node.
Both the exhaustive and pruned canonical algorithms fail identically on these DAGs, confirming that the $\tau$-tuple pruning heuristic is not at fault.

The greedy $\mathrm{D2S}$ handles these DAGs correctly (P1 passes) by encoding the spurious edge via a \texttt{C}/\texttt{c} token after both endpoints are in the CDLL.
The canonical backtracking search, however, tracks remaining edges under the assumption that all edges terminate at operation nodes, and the edge into a \textsc{Var} node violates this invariant.

These failures do not affect Conjecture~\ref{thm:invariant}.
The conjecture requires that both DAGs satisfy the reachability condition of Conjecture~\ref{thm:roundtrip}: ``every non-variable node of $D$ is reachable from some variable via directed paths.''
While the three failing DAGs technically satisfy this condition (the edge into a \textsc{Var} node does not obstruct reachability of operation nodes), they contain a degenerate structure---an input to a leaf node---that lies outside the domain for which $\mathrm{D2S}$ was designed.
A normalization step that strips incoming edges from \textsc{Var} nodes before canonicalization would eliminate these failures entirely.

Figure~\ref{fig:p3_invariance} illustrates canonical invariance and idempotence for $\sin(x_0) + \cos(x_0)$, a well-formed DAG where canonicalization succeeds: the canonical string $w^{**} = \texttt{VcVspv+Ppc}$ is a fixed point of the compose-and-recanonicalize map.

\paragraph{Summary.}
Properties P1, P2, and P4 hold without exception across all $14{,}841$ valid samples.
P3 holds for $100\%$ of the $9{,}125$ non-timeout samples whose DAGs contain no edges into \textsc{Var} nodes.
The three P3 failures are caused by a single degenerate DAG structure---an edge into a \textsc{Var} node---that the canonical $\mathrm{D2S}$ cannot encode.
This structure is semantically vacuous and can be eliminated by a normalization step.
The failures do not contradict Conjectures~\ref{thm:roundtrip} or~\ref{thm:invariant}.

\subsection{Search Space Reduction (P5)}
\label{sec:results_search_space}

We report the results of the controlled permutation experiment described in \S\ref{sec:exp_search_space}.
A total of $961$ structurally unique random DAGs were analysed across $k \in \{1, \ldots, 12\}$ internal nodes and $m \in \{1, 2\}$ input variables.
For $k \leq 8$, all $k!$ permutations were enumerated exhaustively ($721$ DAGs); for $k \in \{9, \ldots, 12\}$, $50{,}000$ random permutations were sampled per DAG ($240$ DAGs).

\begin{figure}[htbp]
\centering
\includegraphics[width=\linewidth]{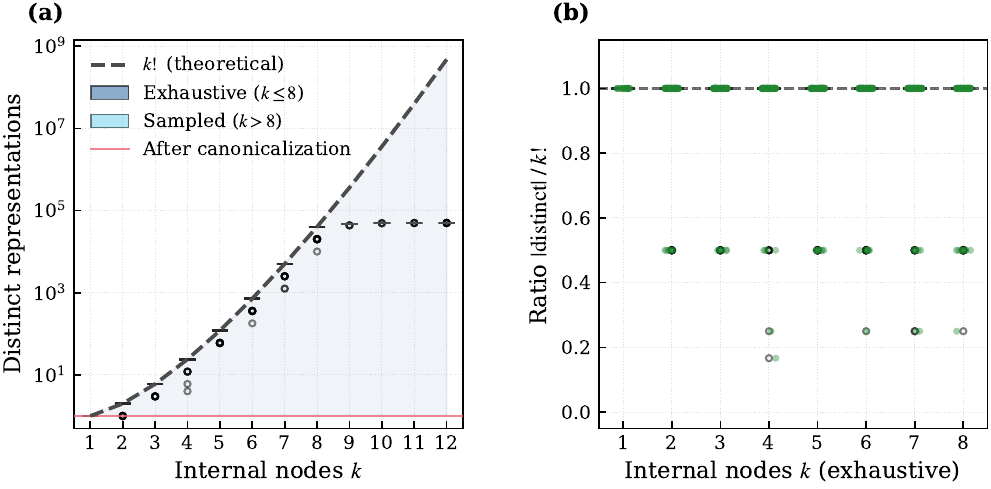}
\caption{Empirical validation of the $\Theta(k!)$ search-space reduction.
For each of $961$ random expression DAGs with $k$ internal nodes ($m \in \{1,2\}$ variables), all $k!$ permutations of internal node IDs were applied (exhaustively for $k \leq 8$; $50{,}000$ sampled for $k > 8$), producing structurally distinct labeled DAGs that encode the same mathematical expression.
\textbf{(a)}~Number of distinct representations (node numberings modulo automorphisms) per expression, compared with the theoretical $k!$ curve (dashed).
For $k \leq 8$, the exhaustive box plots sit on the $k!$ line across five orders of magnitude, confirming that the equivalence class size equals $k!/\lvert\mathrm{Aut}(D)\rvert$.
For $k > 8$, sampled counts are lower bounds.
The red horizontal line at $y = 1$ marks the result after canonicalization: all equivalent representations collapse to a single canonical string (shaded region = eliminated redundancy).
\textbf{(b)}~Normalized ratio of measured distinct representations to $k!$ for the exhaustive range ($k \leq 8$).
The majority of DAGs achieve the full $k!$ bound (ratio $= 1.0$); outliers below $1.0$ correspond to DAGs with non-trivial automorphism groups, where $n_{\text{distinct}} = k!/\lvert\mathrm{Aut}(D)\rvert$ as predicted by the Orbit-Stabilizer theorem.
Canonical invariance was verified at $100\%$ across all $961$ DAGs.}
\label{fig:search_space}
\end{figure}

\paragraph{Distinct representations.}
Figure~\ref{fig:search_space}(a) plots the number of distinct structural fingerprints per DAG against $k$, on a logarithmic scale, alongside the theoretical $k!$ curve.
In the exhaustive range ($k \leq 8$), the box plots sit on the $k!$ line across five orders of magnitude---from $k! = 1$ at $k = 1$ to $k! = 40{,}320$ at $k = 8$---confirming that the equivalence class size equals $k! / \lvert\mathrm{Aut}(D)\rvert$ (Equation~\ref{eq:orbit_stabilizer}).
At $k = 8$, the median DAG admits the full $40{,}320$ distinct representations, confirming the scale of redundancy described in Section~\ref{sec:introduction}.
For $k > 8$, the sampled counts are lower bounds on the true number of distinct representations, yet they continue to track the $k!$ growth rate.
After canonicalization, all equivalent representations collapse to a single canonical string (red line at $y = 1$ in Figure~\ref{fig:search_space}(a)); the shaded region between the data and this line represents the redundancy that canonicalization eliminates.

\paragraph{Automorphism structure.}
Figure~\ref{fig:search_space}(b) shows the normalized ratio $n_{\text{distinct}} / k!$ for the exhaustive range.
The majority of DAGs achieve the full $k!$ bound (ratio $= 1.0$), corresponding to generic expressions with trivial automorphism group ($\lvert\mathrm{Aut}(D)\rvert = 1$).
Outliers appear at ratio $= 0.5$ ($\lvert\mathrm{Aut}(D)\rvert = 2$), $0.25$ ($\lvert\mathrm{Aut}(D)\rvert = 4$), and ${\approx}\,0.17$ ($\lvert\mathrm{Aut}(D)\rvert = 6$).
These correspond to DAGs with non-trivial symmetry---for example, a commutative binary node (\textsc{Add} or \textsc{Mul}) whose two subtrees are isomorphic, yielding $\lvert\mathrm{Aut}(D)\rvert = 2$.
The observed ratios match the predictions of the Orbit-Stabilizer theorem exactly for every exhaustively verified DAG.

\paragraph{Canonical invariance.}
For each DAG, canonical invariance was verified on a subset of permutations by recomputing $\mathrm{canonical}(D_\pi)$ and checking equality with $w^{**}$ (\S\ref{sec:exp_search_space}, step~4).
Across all $961$ DAGs, the success rate is $100\%$: no permutation produced a canonical string different from the original.
No failures or timeouts were observed.

\paragraph{Summary.}
The permutation analysis confirms that each expression with $k$ internal nodes admits $k! / \lvert\mathrm{Aut}(D)\rvert$ structurally distinct node-numbering representations, and that the pruned canonical string collapses all of them to a single form.
The theoretical $\Theta(k!)$ bound is tight for generic DAGs and the reduction is exact: canonicalization eliminates every equivalent representation across the $961$ DAGs tested.

\subsection{Canonicalization Scalability}
\label{sec:results_timing}

The scalability experiment described in \S\ref{sec:exp_timing} has been executed on the Picasso supercomputer.
Results are under analysis and will be reported in a subsequent revision.

\subsection{Metric Space Properties}
\label{sec:results_metric}

If Conjecture~\ref{thm:invariant} holds, the Levenshtein distance~\citep{levenshtein1966} on pruned canonical strings satisfies the identity of indiscernibles and therefore defines a metric on isomorphism classes of expression DAGs.
We probe this metric with two experiments: a shortest-path distance computation between selected expression pairs (\S\ref{sec:results_shortest_path}) and a distance-1 neighbourhood analysis around a single expression (\S\ref{sec:results_neighborhood}).

\subsubsection{Shortest-Path Distance}
\label{sec:results_shortest_path}

Table~\ref{tab:shortest_path} reports the Levenshtein distances between the pruned canonical strings of five expression pairs, selected to span a range of structural relationships (\S\ref{sec:exp_shortest_path}).

\begin{table}[htbp]
\centering
\caption{Levenshtein distances between pruned canonical strings for five expression pairs.
  Each pair is designed to isolate a specific structural relationship.}
\label{tab:shortest_path}
\begin{tabular}{clllc}
\toprule
Pair & Expression 1 & Expression 2 & Relationship & $d_{\mathrm{Lev}}$ \\
\midrule
A & $\sin(x)$             & $x^2 + x$                & Different structure  & 10 \\
B & $\sin(x) + \cos(x)$   & $\sin(x) \cdot \cos(x)$  & Same operands, different combinator & 1 \\
C & $x^2$                 & $x^3 + x^2 + x$          & Containment          & 12 \\
D & $\exp(x)$             & $\log(x)$                & Same topology, different unary op & 1 \\
E & $\sin(x) + y^2$       & $\cos(x) \cdot y$        & Mixed two-variable   & 12 \\
\bottomrule
\end{tabular}
\end{table}

The distances align with the structural similarity between the paired expressions.
Pairs B and D both yield $d_{\mathrm{Lev}} = 1$, corresponding to a single character substitution in the canonical string: the combinator label \texttt{+}~$\to$~\texttt{*} for pair B, and the unary label \texttt{e}~$\to$~\texttt{l} for pair D.
In both cases the DAG topology is identical and only the operation type at one node differs, which the encoding captures as a single-character change.

Pair A ($d_{\mathrm{Lev}} = 10$) and pair C ($d_{\mathrm{Lev}} = 12$) involve expressions that differ in both topology and size, requiring multiple insertions, deletions, and substitutions to transform one canonical string into the other.
Pair C is particularly informative: although $x^2$ is structurally contained in $x^3 + x^2 + x$, the canonical strings of the two expressions share no common prefix, so the edit distance reflects the full structural gap rather than a simple extension.

Pair E ($d_{\mathrm{Lev}} = 12$) combines differences in operator type, DAG topology, and variable usage within a shared two-variable setting ($m = 2$).

Figure~\ref{fig:shortest_path} traces a separate example: the step-by-step Levenshtein path between $\cos(x_0) + x_0$ and $\cos(x_0) + 1$ ($d_{\mathrm{Lev}} = 6$), chosen because the intermediate distance yields a nontrivial yet visually tractable path.
Each intermediate string produces a valid expression DAG.
The path traverses structurally distinct intermediate forms, including a $\cos(\cos(x_0))$ detour, illustrating that the edit distance does not always follow a monotone structural interpolation.

\subsubsection{Distance-1 Neighbourhood}
\label{sec:results_neighborhood}

We analyse the distance-1 neighbourhood of $\sin(x_0) + \cos(x_0)$ (canonical string $w^{**} = \texttt{VcVspv+Ppc}$, length $L = 10$, character-level alphabet size $|\mathcal{A}| = 17$), following the protocol of \S\ref{sec:exp_neighborhood}.
The total neighbourhood consists of $357$ strings: $10$ deletions, $160$ substitutions, and $187$ insertions (Equation~\ref{eq:neighborhood_size}).

Table~\ref{tab:neighbourhood} summarises the outcome by edit operation type.

\begin{table}[htbp]
\centering
\caption{Distance-1 neighbourhood of $\sin(x_0) + \cos(x_0)$ ($w^{**} = \texttt{VcVspv+Ppc}$), disaggregated by edit operation.
  $N_{\mathrm{total}}$: number of Lev-1 strings; $N_{\mathrm{valid}}$: strings that produce a valid non-trivial DAG; $N_{\mathrm{unique}}$: distinct canonical forms among valid strings.}
\label{tab:neighbourhood}
\begin{tabular}{lrrrc}
\toprule
Operation & $N_{\mathrm{total}}$ & $N_{\mathrm{valid}}$ & $N_{\mathrm{unique}}$ & Redundancy \\
\midrule
Deletion     &  10 &   5 &  3 & 40.0\% \\
Substitution & 160 &  60 & 35 & 41.7\% \\
Insertion    & 187 &  67 & 18 & 73.1\% \\
\midrule
All          & 357 & 132 & 38 & 71.2\% \\
\bottomrule
\end{tabular}
\end{table}

Of the $357$ neighbours, $132$ ($37.0\%$) produce valid DAGs with at least one internal node.
The remaining $225$ strings either fail to parse into a valid DAG or yield trivial variable-only graphs.
Among the $132$ valid neighbours, only $38$ produce canonical strings distinct from one another and from the original, yielding an overall redundancy rate of $71.2\%$.
The per-operation unique counts in Table~\ref{tab:neighbourhood} sum to more than $38$ because some canonical forms are reachable by more than one edit type.
An additional $34$ valid strings canonicalize back to the original $w^{**}$, meaning that the corresponding edits are ``absorbed'' by the canonicalization and do not change the represented expression.

The three edit operations exhibit different redundancy profiles.
Deletions have the lowest absolute count ($5$ valid out of $10$) but moderate redundancy ($40.0\%$), because removing a single character from a short string often destroys the DAG structure entirely.
Substitutions produce the most valid neighbours in absolute terms ($60$) and retain moderate redundancy ($41.7\%$): many substitutions change only the label character of a compound token, mapping to a distinct operation type and hence a distinct canonical form.
Insertions generate the most total candidates ($187$) and the highest redundancy ($73.1\%$): many inserted characters produce movement tokens or no-ops that the canonicalization discards, collapsing the resulting DAGs back to a small set of canonical forms.

Figure~\ref{fig:neighbourhood} visualises the neighbourhood, with $d$ labels indicating the Levenshtein distance between each neighbour's canonical string and the original.
Substitution neighbours at $d = 1$ correspond to single-label changes (e.g., $+\to\times$, $+\to-$) that preserve the operand structure, while insertion and deletion neighbours at $d > 1$ produce structurally different DAGs.
The high redundancy rate illustrates the search-space compression that canonicalization provides: nearly three quarters of the valid perturbations collapse to a previously seen canonical form.